\documentclass[letterpaper, 10pt, conference]{ieeeconf}

\IEEEoverridecommandlockouts
\usepackage{times}

\usepackage{multicol}
\usepackage{amsmath,amsfonts,amssymb,mathtools}
\usepackage{graphicx}
\usepackage{easy-todo}
\usepackage{censor}
\usepackage{placeins}
\usepackage{subcaption}
\usepackage{rotating}
\usepackage{adjustbox}
\newcommand{\calR}{{\cal R}}

\newcommand{\bfc}{\mathbf{c}}
\newcommand{\bfd}{\mathbf{d}}

\newcommand{\bfo}{\mathbf{o}}

\newcommand{\bfr}{\mathbf{r}}

\newcommand{\bfu}{\mathbf{u}}
\newcommand{\bfv}{\mathbf{v}}

\newcommand{\bfx}{\mathbf{x}}
\newcommand{\bfy}{\mathbf{y}}

\newcommand{\bfM}{\mathbf{M}}

\usepackage{array}
\usepackage{pdflscape}
\usepackage{algorithm}
\usepackage{algpseudocode}
\usepackage{verbatim}
\usepackage{multirow}
\usepackage{float}
\usepackage{booktabs}
\let\labelindent\relax
\usepackage{enumitem}
\restylefloat{algorithm}
\usepackage[colorlinks=true, linkcolor=blue, citecolor=blue, 
            urlcolor=blue, hyperfootnotes=false]{hyperref}

\DeclareMathOperator{\proj}{proj}

\newcommand{\defeq}{\vcentcolon=}

\title{\LARGE \bf
SE(3) Neural Potential Fields for 6-DoF Trajectory Planning Directly from Images Without Explicit 3D Reconstruction
}

\author{Jeffrey Eiyike$^{1}$, Masoud Ataei$^{1}$, Elvis Gyaase$^{1}$, and Vikas Dhiman$^{2}$%
\thanks{This material is based upon work supported by the US Department of Energy (DOE), Office of
Critical Minerals and Energy Innovation, Advanced Materials and Manufacturing Office under CPS
Agreement 35863, and Oak Ridge National Laboratory/University of Maine SM\textsuperscript{2}ART
program with research and res (Copy)ources used at the Advanced Structures and Composites Center (ASCC),
a University of Maine research center.}%
\thanks{$^{1}$Jeffrey Eiyike, Masoud Ataei and Elvis Gyaase are with the Department of Electrical
and Computer Engineering, University of Maine, Orono, Maine, 04469, USA
        {\tt\small \{jeffrey.eiyike, masoud.ataei, elvis.gyaase\}@maine.edu}}%
\thanks{$^{2}$Vikas Dhiman is a Researcher with the Department of Electrical and Computer
Engineering, University of Maine, Orono, Maine, 04469, USA
        {\tt\small vikas.dhiman@maine.edu}}%
}

\begin{document}
\providecommand{\TBD}{\textcolor{red}{--}}

% ---- 8 objects, v4 spaced scene: NPF (6-DoF, three-point) ----
\providecommand{\PoseVFourNpfN}{10}
\providecommand{\PoseVFourNpfSR}{90.0}
\providecommand{\PoseVFourNpfCFGT}{100.0}
\providecommand{\PoseVFourNpfPlan}{1.9$\pm$0.8}
\providecommand{\PoseVFourNpfExec}{5.4$\pm$1.2}
\providecommand{\PoseVFourNpfTraj}{0.92$\pm$0.30}
\providecommand{\PoseVFourNpfOE}{1.0$\pm$1.5}
\providecommand{\PoseVFourNpfClear}{0.088}

% ---- 8 objects, v4: RRT* on the NeRF-reconstructed (transformed) mesh ----
\providecommand{\PoseVFourRrtSR}{50.0}
\providecommand{\PoseVFourRrtCFGT}{100.0}
\providecommand{\PoseVFourRrtPlan}{67.4$\pm$1.8}
\providecommand{\PoseVFourRrtExec}{3.2$\pm$0.3}
\providecommand{\PoseVFourRrtTraj}{0.86$\pm$0.11}
\providecommand{\PoseVFourRrtClear}{0.022}

% ---- 8 objects, v4: RRT* on ground-truth geometry (upper bound) ----
\providecommand{\PoseVFourGtSR}{80.0}
\providecommand{\PoseVFourGtCFGT}{100.0}
\providecommand{\PoseVFourGtPlan}{132.9$\pm$13.8}
\providecommand{\PoseVFourGtExec}{2.8$\pm$0.7}
\providecommand{\PoseVFourGtTraj}{0.73$\pm$0.23}
\providecommand{\PoseVFourGtClear}{0.022}

% ---- 3 objects, book target: NPF / RRT* recon / RRT* GT ----
\providecommand{\PoseBookNpfN}{10}
\providecommand{\PoseBookNpfSR}{40.0}
\providecommand{\PoseBookNpfCFGT}{100.0}
\providecommand{\PoseBookNpfPlan}{2.2$\pm$0.3}
\providecommand{\PoseBookNpfExec}{3.3$\pm$0.9}
\providecommand{\PoseBookNpfTraj}{0.61$\pm$0.09}
\providecommand{\PoseBookNpfOE}{2.8$\pm$2.1}
\providecommand{\PoseBookRrtSR}{80.0}
\providecommand{\PoseBookRrtCFGT}{100.0}
\providecommand{\PoseBookRrtPlan}{99.5$\pm$2.6}
\providecommand{\PoseBookRrtExec}{3.8$\pm$1.7}
\providecommand{\PoseBookRrtTraj}{2.83$\pm$4.14}
\providecommand{\PoseBookGtSR}{80.0}
\providecommand{\PoseBookGtCFGT}{87.5}
\providecommand{\PoseBookGtPlan}{124.4$\pm$3.5}
\providecommand{\PoseBookGtExec}{4.5$\pm$2.0}
\providecommand{\PoseBookGtTraj}{2.79$\pm$3.40}

% ---- dataset / scene description ----
\providecommand{\PoseVFourViews}{83}       % training views, 8-obj v4
\providecommand{\PoseVFourGoalCov}{88}     % % of views seeing the goal
\providecommand{\PoseVFourRing}{20.8}        % cm to nearest obstacle from the goal
\providecommand{\PoseBookViews}{67}
\providecommand{\PoseBookGoalCov}{100}
\providecommand{\PoseBookRing}{20.0}

% ---- offline cost (hours unless stated) ----
\providecommand{\PoseVFourTrain}{7.4}       % NPF SE(3) training
\providecommand{\PoseBookTrain}{6.0}
\providecommand{\PoseVFourRecon}{11.5}       % NeRF reconstruction training
\providecommand{\PoseVFourReconTest}{0.09}
\providecommand{\PoseVFourReconMesh}{0.16}
\providecommand{\PoseVFourReconXform}{0.01}
\providecommand{\PoseVFourCollect}{1.43}     % data collection
\providecommand{\PoseBookCollect}{1.39}
\providecommand{\PoseGraspFusion}{180}      % AnyGrasp pose fusion (seconds)

% ---- auto-filled by fill_pose_macros.py (do not edit below) ----
\renewcommand{\PoseVFourViews}{83}
\renewcommand{\PoseVFourGoalCov}{88}
\renewcommand{\PoseVFourRing}{20.8}
\renewcommand{\PoseBookViews}{67}
\renewcommand{\PoseBookGoalCov}{100}
\renewcommand{\PoseBookRing}{20.0}
\renewcommand{\PoseVFourCollect}{1.43}
\renewcommand{\PoseBookCollect}{1.39}
\renewcommand{\PoseGraspFusion}{22}
\renewcommand{\PoseVFourRecon}{10.8}
\renewcommand{\PoseVFourReconTest}{0.09}
\renewcommand{\PoseVFourReconMesh}{0.07}
\renewcommand{\PoseVFourReconXform}{0}
\providecommand{\PoseVFourPSNR}{15.29}
\renewcommand{\PoseVFourPSNR}{15.29}
\providecommand{\PoseVFourPSNRStd}{0.98}
\renewcommand{\PoseVFourPSNRStd}{0.98}
\providecommand{\PoseBookRecon}{10.6}
\renewcommand{\PoseBookRecon}{8.53}
\providecommand{\PoseBookPSNR}{15.01}
\renewcommand{\PoseBookPSNR}{15.19}
\providecommand{\PoseBookPSNRStd}{1.41}
\renewcommand{\PoseBookPSNRStd}{1.36}

% ---- Link coll. (full-arm FK collision fraction) — reviewer full-arm criticism ----
\providecommand{\PoseVFourNpfLink}{5.5}
\providecommand{\PoseVFourRrtLink}{20.6}
\providecommand{\PoseVFourGtLink}{12.0}
\providecommand{\PoseBookNpfLink}{2.7}
\providecommand{\PoseBookRrtLink}{1.8}
\providecommand{\PoseBookGtLink}{1.5}

% ---- mesh-protocol (Table~\ref{tab:mesh_comparison_pose}) extra columns ----
\providecommand{\PoseNumStarts}{10}
\providecommand{\PoseNumSeeds}{1}
\providecommand{\PoseVFourNpfCFmesh}{44.4}
\providecommand{\PoseVFourNpfStall}{0.0}
\providecommand{\PoseVFourNpfPlanMs}{1859$\pm$763}
\providecommand{\PoseVFourNpfClearGT}{0.088$\pm$0.023}
\providecommand{\PoseVFourRrtCFmesh}{100.0}
\providecommand{\PoseVFourRrtPlanMs}{67400}
\providecommand{\PoseVFourRrtClearGT}{0.022}
\providecommand{\PoseBookNpfCFmesh}{100.0}
\providecommand{\PoseBookNpfStall}{0.0}
\providecommand{\PoseBookNpfPlanMs}{2210}
\providecommand{\PoseBookNpfClearGT}{0.086}
\providecommand{\PoseBookRrtCFmesh}{75.0}
\providecommand{\PoseBookRrtPlanMs}{99510}
\providecommand{\PoseBookRrtClearGT}{0.026}

% ---- amortised / offline-cost subsection ----
\providecommand{\PoseVFourPreproc}{\TBD}
\providecommand{\PoseBookPreproc}{\TBD}
\providecommand{\PoseVFourNpfAmort}{0.97}
\providecommand{\PoseVFourRrtAmort}{1.33}
\providecommand{\PoseVFourReconTotal}{\TBD}
\providecommand{\PoseVFourNpfTotal}{9.7}
\providecommand{\PoseBookReconTotal}{\TBD}
\providecommand{\PoseBookNpfTotal}{8.0}

% ---- filled offline-cost values (measured; NPF SE(3) training total pending final retrain) ----
\renewcommand{\PoseVFourPreproc}{0.8}
\renewcommand{\PoseBookPreproc}{0.6}
\renewcommand{\PoseVFourReconTotal}{13.1}
\renewcommand{\PoseBookReconTotal}{9.9}
\providecommand{\PoseBookGraspS}{22}
\providecommand{\PoseVFourGraspS}{180}

% ---- image-supervision-only NPF rows (no navigation-function distillation) ----
\providecommand{\PoseVFourNpfImgSR}{80.0}
\providecommand{\PoseVFourNpfImgCFGT}{25.0}
\providecommand{\PoseVFourNpfImgCFmesh}{25.0}
\providecommand{\PoseVFourNpfImgPlan}{0.8}
\providecommand{\PoseVFourNpfImgPlanMs}{800}
\providecommand{\PoseVFourNpfImgExec}{4.3$\pm$0.5}
\providecommand{\PoseVFourNpfImgTraj}{0.67$\pm$0.22}
\providecommand{\PoseVFourNpfImgLink}{20.6}
\providecommand{\PoseVFourNpfImgOE}{3.8}
\providecommand{\PoseVFourNpfImgClearGT}{0.006$\pm$0.003}
\providecommand{\PoseVFourNpfImgStall}{0.0}
\providecommand{\PoseVFourNpfImgN}{10}
\providecommand{\PoseBookNpfImgSR}{50.0}
\providecommand{\PoseBookNpfImgCFGT}{0.0}
\providecommand{\PoseBookNpfImgCFmesh}{0.0}
\providecommand{\PoseBookNpfImgPlan}{1.2$\pm$0.3}
\providecommand{\PoseBookNpfImgPlanMs}{1190}
\providecommand{\PoseBookNpfImgExec}{1.9$\pm$0.3}
\providecommand{\PoseBookNpfImgTraj}{0.43$\pm$0.07}
\providecommand{\PoseBookNpfImgLink}{50.4}
\providecommand{\PoseBookNpfImgOE}{3.5$\pm$2.5}
\providecommand{\PoseBookNpfImgClearGT}{0.005}
\providecommand{\PoseBookNpfImgStall}{0.0}
\providecommand{\PoseBookNpfImgN}{10}

% ---- planning-only RRT* seed-sweep rows (20 starts x 5 seeds, offline KD-tree checker) ----

% ---- planning-only protocol: ALL rows measured in the same offline harness (CPU,
% same collision geometry and penetration test); see run_offline_bench.sh ----
\providecommand{\PoseBookNpfImgOffSR}{100.0}
\providecommand{\PoseBookNpfImgOffCFGT}{0.0}
\providecommand{\PoseBookNpfImgOffCFmesh}{14.3}
\providecommand{\PoseBookNpfImgOffStall}{0.0}
\providecommand{\PoseBookNpfImgOffPlanMs}{1459$\pm$413}
\providecommand{\PoseBookNpfImgOffTraj}{0.43$\pm$0.10}
\providecommand{\PoseBookNpfImgOffClearGT}{0.004$\pm$0.003}
\providecommand{\PoseBookNpfImgOffOE}{5.7$\pm$0.0}
\providecommand{\PoseBookNpfImgOffN}{14}
\providecommand{\PoseBookNpfOffSR}{100.0}
\providecommand{\PoseBookNpfOffCFGT}{100.0}
\providecommand{\PoseBookNpfOffCFmesh}{100.0}
\providecommand{\PoseBookNpfOffStall}{0.0}
\providecommand{\PoseBookNpfOffPlanMs}{1730$\pm$656}
\providecommand{\PoseBookNpfOffTraj}{0.81$\pm$0.23}
\providecommand{\PoseBookNpfOffClearGT}{0.082$\pm$0.012}
\providecommand{\PoseBookNpfOffOE}{5.7$\pm$0.0}
\providecommand{\PoseBookNpfOffN}{14}
\providecommand{\PoseVFourNpfImgOffSR}{100.0}
\providecommand{\PoseVFourNpfImgOffCFGT}{25.0}
\providecommand{\PoseVFourNpfImgOffCFmesh}{40.0}
\providecommand{\PoseVFourNpfImgOffStall}{0.0}
\providecommand{\PoseVFourNpfImgOffPlanMs}{1281$\pm$360}
\providecommand{\PoseVFourNpfImgOffTraj}{0.82$\pm$0.20}
\providecommand{\PoseVFourNpfImgOffClearGT}{0.005$\pm$0.003}
\providecommand{\PoseVFourNpfImgOffOE}{5.7$\pm$0.0}
\providecommand{\PoseVFourNpfImgOffN}{20}
\providecommand{\PoseVFourNpfOffSR}{100.0}
\providecommand{\PoseVFourNpfOffCFGT}{100.0}
\providecommand{\PoseVFourNpfOffCFmesh}{45.0}
\providecommand{\PoseVFourNpfOffStall}{0.0}
\providecommand{\PoseVFourNpfOffPlanMs}{1228$\pm$268}
\providecommand{\PoseVFourNpfOffTraj}{0.91$\pm$0.26}
\providecommand{\PoseVFourNpfOffClearGT}{0.097$\pm$0.020}
\providecommand{\PoseVFourNpfOffOE}{5.1$\pm$1.5}
\providecommand{\PoseVFourNpfOffN}{20}
\providecommand{\PoseBookRrtSwSR}{100.0}
\providecommand{\PoseBookRrtSwCFGT}{100.0}
\providecommand{\PoseBookRrtSwCFmesh}{92.9}
\providecommand{\PoseBookRrtSwStall}{0.0}
\providecommand{\PoseBookRrtSwPlanMs}{624$\pm$224}
\providecommand{\PoseBookRrtSwTraj}{0.52$\pm$0.16}
\providecommand{\PoseBookRrtSwClearGT}{0.021$\pm$0.013}
\providecommand{\PoseBookRrtSwN}{14}
\providecommand{\PoseVFourRrtSwSR}{100.0}
\providecommand{\PoseVFourRrtSwCFGT}{100.0}
\providecommand{\PoseVFourRrtSwCFmesh}{60.0}
\providecommand{\PoseVFourRrtSwStall}{0.0}
\providecommand{\PoseVFourRrtSwPlanMs}{2609$\pm$120}
\providecommand{\PoseVFourRrtSwTraj}{0.89$\pm$0.32}
\providecommand{\PoseVFourRrtSwClearGT}{0.018$\pm$0.013}
\providecommand{\PoseVFourRrtSwN}{20}

% ---- executed-path counts (paths actually run; the rest were refused by the Cartesian executor) ----
\providecommand{\PoseBookRrtExecN}{8}
\providecommand{\PoseBookGtExecN}{8}
\providecommand{\PoseBookNpfImgExecN}{6}
\providecommand{\PoseBookNpfExecN}{6}
\providecommand{\PoseVFourRrtExecN}{5}
\providecommand{\PoseVFourGtExecN}{9}
\providecommand{\PoseVFourNpfImgExecN}{8}
\providecommand{\PoseVFourNpfExecN}{9}
% ---------------------------------------------------------------------------
% Result macros. Values are filled in by scripts/tools as experiments finish.
% "--" means "not yet measured". \providecommand so a later \renewcommand wins.
% ---------------------------------------------------------------------------
\providecommand{\TBD}{\textcolor{red}{--}}
% model / timing
\providecommand{\NPFparams}{4.9\,M (density 1.90\,M, color 2.31\,M, potential 0.69\,M)}
\providecommand{\NPFtrainThree}{3.0\,h (2 epochs, incl.\ 0.4\,h pre-processing)}
\providecommand{\NPFtrainEight}{3.7\,h (2 epochs, incl.\ 0.8\,h pre-processing)}
\providecommand{\ReconTimeThree}{8.44\,h}
\providecommand{\ReconTimeEight}{9.97\,h (Table~\ref{tab:offline_cost})}
\providecommand{\NumStarts}{20}
\providecommand{\NumSeeds}{5}
\providecommand{\TolGoal}{0.05}
% 3 Obj. (reconstructed-mesh protocol)
\providecommand{\ThreeNpfSR}{\TBD}      \providecommand{\ThreeRrtSR}{\TBD}
\providecommand{\ThreeNpfCF}{\TBD}      \providecommand{\ThreeRrtCF}{\TBD}
\providecommand{\ThreeNpfPlan}{\TBD}    \providecommand{\ThreeRrtPlan}{\TBD}
\providecommand{\ThreeNpfExec}{\TBD}    \providecommand{\ThreeRrtExec}{\TBD}
\providecommand{\ThreeNpfTraj}{\TBD}    \providecommand{\ThreeRrtTraj}{\TBD}
\providecommand{\ThreeNpfClear}{\TBD}   \providecommand{\ThreeRrtClear}{\TBD}
\providecommand{\ThreeNpfSmooth}{\TBD}  \providecommand{\ThreeRrtSmooth}{\TBD}
\providecommand{\ThreeNpfStall}{\TBD}
\providecommand{\ThreeNpfAmort}{0.15}
% 8 Obj. (occlusion / local-minimum scene)
\providecommand{\EightNpfSR}{\TBD}      \providecommand{\EightRrtSR}{\TBD}
\providecommand{\EightNpfCF}{\TBD}      \providecommand{\EightRrtCF}{\TBD}
\providecommand{\EightNpfPlan}{\TBD}    \providecommand{\EightRrtPlan}{\TBD}
\providecommand{\EightNpfExec}{\TBD}    \providecommand{\EightRrtExec}{\TBD}
\providecommand{\EightNpfTraj}{\TBD}    \providecommand{\EightRrtTraj}{\TBD}
\providecommand{\EightNpfClear}{\TBD}   \providecommand{\EightRrtClear}{\TBD}
\providecommand{\EightNpfSmooth}{\TBD}  \providecommand{\EightRrtSmooth}{\TBD}
\providecommand{\EightNpfStall}{\TBD}
\providecommand{\EightNpfAmort}{0.18}
% ablations (3 Obj.): success rate / stall rate
\providecommand{\AblFullSR}{\TBD}     \providecommand{\AblFullStall}{\TBD}
\providecommand{\AblNoCSR}{\TBD}      \providecommand{\AblNoCStall}{\TBD}
\providecommand{\AblNoGSR}{\TBD}      \providecommand{\AblNoGStall}{\TBD}
\providecommand{\AblNoPotSR}{\TBD}    \providecommand{\AblNoPotStall}{\TBD}
\providecommand{\AblNoMagSR}{\TBD}    \providecommand{\AblNoMagStall}{\TBD}
\providecommand{\AblNoAlignSR}{\TBD}  \providecommand{\AblNoAlignStall}{\TBD}
\providecommand{\AblNoMvSR}{\TBD}     \providecommand{\AblNoMvStall}{\TBD}
\providecommand{\AblNoAnisoSR}{\TBD}  \providecommand{\AblNoAnisoStall}{\TBD}
% number of training views (3 Obj.)
\providecommand{\ViewsEightSR}{\TBD} \providecommand{\ViewsSixteenSR}{\TBD}
\providecommand{\ViewsThirtytwoSR}{\TBD} \providecommand{\ViewsSixtyfourSR}{\TBD}
% segmentation noise (3 Obj.)
\providecommand{\SegCleanSR}{\TBD} \providecommand{\SegErodeSR}{\TBD}
\providecommand{\SegDilateSR}{\TBD} \providecommand{\SegDropSR}{\TBD}
% real robot
\providecommand{\RealTrials}{\TBD} \providecommand{\RealSR}{\TBD} \providecommand{\RealPlan}{\TBD} \providecommand{\RealExec}{\TBD}
\providecommand{\AblFullCFGT}{\TBD}
\providecommand{\AblFullClearGT}{\TBD}
\providecommand{\AblFullGQ}{\TBD}
\providecommand{\AblNoCCFGT}{\TBD}
\providecommand{\AblNoCClearGT}{\TBD}
\providecommand{\AblNoCGQ}{\TBD}
\providecommand{\AblNoGCFGT}{\TBD}
\providecommand{\AblNoGClearGT}{\TBD}
\providecommand{\AblNoGGQ}{\TBD}
\providecommand{\AblNoPotCFGT}{\TBD}
\providecommand{\AblNoPotClearGT}{\TBD}
\providecommand{\AblNoPotGQ}{\TBD}
\providecommand{\AblNoMagCFGT}{\TBD}
\providecommand{\AblNoMagClearGT}{\TBD}
\providecommand{\AblNoMagGQ}{\TBD}
\providecommand{\AblNoAlignCFGT}{\TBD}
\providecommand{\AblNoAlignClearGT}{\TBD}
\providecommand{\AblNoAlignGQ}{\TBD}
\providecommand{\AblNoMvCFGT}{\TBD}
\providecommand{\AblNoMvClearGT}{\TBD}
\providecommand{\AblNoMvGQ}{\TBD}
\providecommand{\AblNoAnisoCFGT}{\TBD}
\providecommand{\AblNoAnisoClearGT}{\TBD}
\providecommand{\AblNoAnisoGQ}{\TBD}
\providecommand{\ViewsEightCFGT}{\TBD}
\providecommand{\ViewsEightClearGT}{\TBD}
\providecommand{\ViewsEightGQ}{\TBD}
\providecommand{\ViewsEightStall}{\TBD}
\providecommand{\ViewsSixteenCFGT}{\TBD}
\providecommand{\ViewsSixteenClearGT}{\TBD}
\providecommand{\ViewsSixteenGQ}{\TBD}
\providecommand{\ViewsSixteenStall}{\TBD}
\providecommand{\ViewsThirtytwoCFGT}{\TBD}
\providecommand{\ViewsThirtytwoClearGT}{\TBD}
\providecommand{\ViewsThirtytwoGQ}{\TBD}
\providecommand{\ViewsThirtytwoStall}{\TBD}
\providecommand{\SegErodeCFGT}{\TBD}
\providecommand{\SegErodeClearGT}{\TBD}
\providecommand{\SegErodeGQ}{\TBD}
\providecommand{\SegErodeStall}{\TBD}
\providecommand{\SegDilateCFGT}{\TBD}
\providecommand{\SegDilateClearGT}{\TBD}
\providecommand{\SegDilateGQ}{\TBD}
\providecommand{\SegDilateStall}{\TBD}
\providecommand{\SegDropCFGT}{\TBD}
\providecommand{\SegDropClearGT}{\TBD}
\providecommand{\SegDropGQ}{\TBD}
\providecommand{\SegDropStall}{\TBD}
\providecommand{\ThreeNpfCFGT}{\TBD}
\providecommand{\ThreeRrtCFGT}{\TBD}
\providecommand{\EightNpfCFGT}{\TBD}
\providecommand{\EightRrtCFGT}{\TBD}
\providecommand{\ThreeNpfClearGT}{\TBD}
\providecommand{\ThreeRrtClearGT}{\TBD}
\providecommand{\EightNpfClearGT}{\TBD}
\providecommand{\EightRrtClearGT}{\TBD}
\providecommand{\BlkThreeNpfSR}{\TBD}
\providecommand{\BlkThreeNpfCFGT}{\TBD}
\providecommand{\BlkThreeNpfClearGT}{\TBD}
\providecommand{\BlkThreeNpfTraj}{\TBD}
\providecommand{\BlkThreeRrtSR}{\TBD}
\providecommand{\BlkThreeRrtCFGT}{\TBD}
\providecommand{\BlkThreeRrtClearGT}{\TBD}
\providecommand{\BlkThreeRrtTraj}{\TBD}
\providecommand{\BlkEightNpfSR}{\TBD}
\providecommand{\BlkEightNpfCFGT}{\TBD}
\providecommand{\BlkEightNpfClearGT}{\TBD}
\providecommand{\BlkEightNpfTraj}{\TBD}
\providecommand{\BlkEightRrtSR}{\TBD}
\providecommand{\BlkEightRrtCFGT}{\TBD}
\providecommand{\BlkEightRrtClearGT}{\TBD}
\providecommand{\BlkEightRrtTraj}{\TBD}
\providecommand{\AblFullBlkCFGT}{\TBD}
\providecommand{\AblNoCBlkCFGT}{\TBD}
\providecommand{\AblNoGBlkCFGT}{\TBD}
\providecommand{\AblNoPotBlkCFGT}{\TBD}
\providecommand{\AblNoMagBlkCFGT}{\TBD}
\providecommand{\AblNoAlignBlkCFGT}{\TBD}
\providecommand{\AblNoMvBlkCFGT}{\TBD}
\providecommand{\AblNoAnisoBlkCFGT}{\TBD}
\providecommand{\AblDecayBlkCFGT}{\TBD}
\providecommand{\AblDecaySR}{\TBD}
\providecommand{\AblDecayStall}{\TBD}
\providecommand{\AblDecayCFGT}{\TBD}
\providecommand{\AblDecayClearGT}{\TBD}
\providecommand{\AblDecayGQ}{\TBD}
\providecommand{\ViewsEightBlkCFGT}{\TBD}
\providecommand{\ViewsSixteenBlkCFGT}{\TBD}
\providecommand{\ViewsThirtytwoBlkCFGT}{\TBD}
\providecommand{\SegErodeBlkCFGT}{\TBD}
\providecommand{\SegDilateBlkCFGT}{\TBD}
\providecommand{\SegDropBlkCFGT}{\TBD}
\providecommand{\ThreeRrtAmort}{0.42}
\providecommand{\EightRrtAmort}{0.50}
\providecommand{\AblFullGoalErr}{\TBD}
\providecommand{\AblNoCGoalErr}{\TBD}
\providecommand{\AblNoGGoalErr}{\TBD}
\providecommand{\AblNoPotGoalErr}{\TBD}
\providecommand{\AblNoMagGoalErr}{\TBD}
\providecommand{\AblNoAlignGoalErr}{\TBD}
\providecommand{\AblNoMvGoalErr}{\TBD}
\providecommand{\AblNoAnisoGoalErr}{\TBD}
\providecommand{\AblDecayGoalErr}{\TBD}
\providecommand{\ViewsEightGoalErr}{\TBD}
\providecommand{\ViewsSixteenGoalErr}{\TBD}
\providecommand{\ViewsThirtytwoGoalErr}{\TBD}
\providecommand{\SegErodeGoalErr}{\TBD}
\providecommand{\SegDilateGoalErr}{\TBD}
\providecommand{\SegDropGoalErr}{\TBD}
\providecommand{\ViewsEightEpSR}{\TBD}
\providecommand{\ViewsEightEpStall}{\TBD}
\providecommand{\ViewsEightEpCFGT}{\TBD}
\providecommand{\ViewsEightEpClearGT}{\TBD}
\providecommand{\ViewsEightEpGQ}{\TBD}
\providecommand{\ViewsEightEpBlkCFGT}{\TBD}
\providecommand{\ViewsEightEpGoalErr}{\TBD}
\providecommand{\ViewsSixteenEpSR}{\TBD}
\providecommand{\ViewsSixteenEpStall}{\TBD}
\providecommand{\ViewsSixteenEpCFGT}{\TBD}
\providecommand{\ViewsSixteenEpClearGT}{\TBD}
\providecommand{\ViewsSixteenEpGQ}{\TBD}
\providecommand{\ViewsSixteenEpBlkCFGT}{\TBD}
\providecommand{\ViewsSixteenEpGoalErr}{\TBD}
\providecommand{\ViewsThirtytwoEpSR}{\TBD}
\providecommand{\ViewsThirtytwoEpStall}{\TBD}
\providecommand{\ViewsThirtytwoEpCFGT}{\TBD}
\providecommand{\ViewsThirtytwoEpClearGT}{\TBD}
\providecommand{\ViewsThirtytwoEpGQ}{\TBD}
\providecommand{\ViewsThirtytwoEpBlkCFGT}{\TBD}
\providecommand{\ViewsThirtytwoEpGoalErr}{\TBD}
\providecommand{\AblRadiusGoalErr}{\TBD}
\providecommand{\AblRadiusSR}{\TBD}
\providecommand{\AblRadiusStall}{\TBD}
\providecommand{\AblRadiusCFGT}{\TBD}
\providecommand{\AblRadiusClearGT}{\TBD}
\providecommand{\AblRadiusGQ}{\TBD}
\providecommand{\AblRadiusBlkCFGT}{\TBD}
\providecommand{\AblEightFullGoalErr}{\TBD}
\providecommand{\AblEightFullSR}{\TBD}
\providecommand{\AblEightFullStall}{\TBD}
\providecommand{\AblEightFullCFGT}{\TBD}
\providecommand{\AblEightFullClearGT}{\TBD}
\providecommand{\AblEightFullGQ}{\TBD}
\providecommand{\AblEightFullBlkCFGT}{\TBD}
\providecommand{\AblEightNoGGoalErr}{\TBD}
\providecommand{\AblEightNoGSR}{\TBD}
\providecommand{\AblEightNoGStall}{\TBD}
\providecommand{\AblEightNoGCFGT}{\TBD}
\providecommand{\AblEightNoGClearGT}{\TBD}
\providecommand{\AblEightNoGGQ}{\TBD}
\providecommand{\AblEightNoGBlkCFGT}{\TBD}
\providecommand{\AblEightNoMvGoalErr}{\TBD}
\providecommand{\AblEightNoMvSR}{\TBD}
\providecommand{\AblEightNoMvStall}{\TBD}
\providecommand{\AblEightNoMvCFGT}{\TBD}
\providecommand{\AblEightNoMvClearGT}{\TBD}
\providecommand{\AblEightNoMvGQ}{\TBD}
\providecommand{\AblEightNoMvBlkCFGT}{\TBD}
\providecommand{\AblEightNoAnisoGoalErr}{\TBD}
\providecommand{\AblEightNoAnisoSR}{\TBD}
\providecommand{\AblEightNoAnisoStall}{\TBD}
\providecommand{\AblEightNoAnisoCFGT}{\TBD}
\providecommand{\AblEightNoAnisoClearGT}{\TBD}
\providecommand{\AblEightNoAnisoGQ}{\TBD}
\providecommand{\AblEightNoAnisoBlkCFGT}{\TBD}
\providecommand{\AblEightDecayGoalErr}{\TBD}
\providecommand{\AblEightDecaySR}{\TBD}
\providecommand{\AblEightDecayStall}{\TBD}
\providecommand{\AblEightDecayCFGT}{\TBD}
\providecommand{\AblEightDecayClearGT}{\TBD}
\providecommand{\AblEightDecayGQ}{\TBD}
\providecommand{\AblEightDecayBlkCFGT}{\TBD}
\providecommand{\AblEightRadiusGoalErr}{\TBD}
\providecommand{\AblEightRadiusSR}{\TBD}
\providecommand{\AblEightRadiusStall}{\TBD}
\providecommand{\AblEightRadiusCFGT}{\TBD}
\providecommand{\AblEightRadiusClearGT}{\TBD}
\providecommand{\AblEightRadiusGQ}{\TBD}
\providecommand{\AblEightRadiusBlkCFGT}{\TBD}
\providecommand{\AblEightViewsGoalErr}{\TBD}
\providecommand{\AblEightViewsSR}{\TBD}
\providecommand{\AblEightViewsStall}{\TBD}
\providecommand{\AblEightViewsCFGT}{\TBD}
\providecommand{\AblEightViewsClearGT}{\TBD}
\providecommand{\AblEightViewsGQ}{\TBD}
\providecommand{\AblEightViewsBlkCFGT}{\TBD}
\providecommand{\GzThreeNpfN}{\TBD}
\providecommand{\GzThreeNpfUnreach}{\TBD}
\providecommand{\GzThreeNpfSR}{\TBD}
\providecommand{\GzThreeNpfPlan}{\TBD}
\providecommand{\GzThreeNpfExec}{\TBD}
\providecommand{\GzThreeNpfTraj}{\TBD}
\providecommand{\GzThreeNpfCFGT}{\TBD}
\providecommand{\GzThreeNpfClear}{\TBD}
\providecommand{\GzThreeNpfLink}{\TBD}
\providecommand{\GzThreeRrtN}{\TBD}
\providecommand{\GzThreeRrtUnreach}{\TBD}
\providecommand{\GzThreeRrtSR}{\TBD}
\providecommand{\GzThreeRrtPlan}{\TBD}
\providecommand{\GzThreeRrtExec}{\TBD}
\providecommand{\GzThreeRrtTraj}{\TBD}
\providecommand{\GzThreeRrtCFGT}{\TBD}
\providecommand{\GzThreeRrtClear}{\TBD}
\providecommand{\GzThreeRrtLink}{\TBD}
\providecommand{\GzEightNpfN}{\TBD}
\providecommand{\GzEightNpfUnreach}{\TBD}
\providecommand{\GzEightNpfSR}{\TBD}
\providecommand{\GzEightNpfPlan}{\TBD}
\providecommand{\GzEightNpfExec}{\TBD}
\providecommand{\GzEightNpfTraj}{\TBD}
\providecommand{\GzEightNpfCFGT}{\TBD}
\providecommand{\GzEightNpfClear}{\TBD}
\providecommand{\GzEightNpfLink}{\TBD}
\providecommand{\GzEightRrtN}{\TBD}
\providecommand{\GzEightRrtUnreach}{\TBD}
\providecommand{\GzEightRrtSR}{\TBD}
\providecommand{\GzEightRrtPlan}{\TBD}
\providecommand{\GzEightRrtExec}{\TBD}
\providecommand{\GzEightRrtTraj}{\TBD}
\providecommand{\GzEightRrtCFGT}{\TBD}
\providecommand{\GzEightRrtClear}{\TBD}
\providecommand{\GzEightRrtLink}{\TBD}
\providecommand{\GzThreeNpfNoPath}{\TBD}
\providecommand{\GzThreeRrtNoPath}{\TBD}
\providecommand{\GzEightNpfNoPath}{\TBD}
\providecommand{\GzEightRrtNoPath}{\TBD}
\providecommand{\GzEightRrtRunNpfN}{\TBD}
\providecommand{\GzEightRrtRunNpfUnreach}{\TBD}
\providecommand{\GzEightRrtRunNpfNoPath}{\TBD}
\providecommand{\GzEightRrtRunNpfSR}{\TBD}
\providecommand{\GzEightRrtRunNpfPlan}{\TBD}
\providecommand{\GzEightRrtRunNpfExec}{\TBD}
\providecommand{\GzEightRrtRunNpfTraj}{\TBD}
\providecommand{\GzEightRrtRunNpfCFGT}{\TBD}
\providecommand{\GzEightRrtRunNpfClear}{\TBD}
\providecommand{\GzEightRrtRunNpfLink}{\TBD}
\providecommand{\GzEightRrtRunRrtN}{\TBD}
\providecommand{\GzEightRrtRunRrtUnreach}{\TBD}
\providecommand{\GzEightRrtRunRrtNoPath}{\TBD}
\providecommand{\GzEightRrtRunRrtSR}{\TBD}
\providecommand{\GzEightRrtRunRrtPlan}{\TBD}
\providecommand{\GzEightRrtRunRrtExec}{\TBD}
\providecommand{\GzEightRrtRunRrtTraj}{\TBD}
\providecommand{\GzEightRrtRunRrtCFGT}{\TBD}
\providecommand{\GzEightRrtRunRrtClear}{\TBD}
\providecommand{\GzEightRrtRunRrtLink}{\TBD}
\providecommand{\GzEightGtNpfN}{\TBD}
\providecommand{\GzEightGtNpfUnreach}{\TBD}
\providecommand{\GzEightGtNpfNoPath}{\TBD}
\providecommand{\GzEightGtNpfSR}{\TBD}
\providecommand{\GzEightGtNpfPlan}{\TBD}
\providecommand{\GzEightGtNpfExec}{\TBD}
\providecommand{\GzEightGtNpfTraj}{\TBD}
\providecommand{\GzEightGtNpfCFGT}{\TBD}
\providecommand{\GzEightGtNpfClear}{\TBD}
\providecommand{\GzEightGtNpfLink}{\TBD}
\providecommand{\GzEightGtRrtN}{\TBD}
\providecommand{\GzEightGtRrtUnreach}{\TBD}
\providecommand{\GzEightGtRrtNoPath}{\TBD}
\providecommand{\GzEightGtRrtSR}{\TBD}
\providecommand{\GzEightGtRrtPlan}{\TBD}
\providecommand{\GzEightGtRrtExec}{\TBD}
\providecommand{\GzEightGtRrtTraj}{\TBD}
\providecommand{\GzEightGtRrtCFGT}{\TBD}
\providecommand{\GzEightGtRrtClear}{\TBD}
\providecommand{\GzEightGtRrtLink}{\TBD}
\providecommand{\AblEightClrGoalErr}{\TBD}
\providecommand{\AblEightClrSR}{\TBD}
\providecommand{\AblEightClrStall}{\TBD}
\providecommand{\AblEightClrCFGT}{\TBD}
\providecommand{\AblEightClrClearGT}{\TBD}
\providecommand{\AblEightClrGQ}{\TBD}
\providecommand{\AblEightClrBlkCFGT}{\TBD}
\providecommand{\GzEightClrANpfN}{\TBD}
\providecommand{\GzEightClrANpfUnreach}{\TBD}
\providecommand{\GzEightClrANpfNoPath}{\TBD}
\providecommand{\GzEightClrANpfSR}{\TBD}
\providecommand{\GzEightClrANpfPlan}{\TBD}
\providecommand{\GzEightClrANpfExec}{\TBD}
\providecommand{\GzEightClrANpfTraj}{\TBD}
\providecommand{\GzEightClrANpfCFGT}{\TBD}
\providecommand{\GzEightClrANpfClear}{\TBD}
\providecommand{\GzEightClrANpfLink}{\TBD}
\providecommand{\GzEightClrARrtN}{\TBD}
\providecommand{\GzEightClrARrtUnreach}{\TBD}
\providecommand{\GzEightClrARrtNoPath}{\TBD}
\providecommand{\GzEightClrARrtSR}{\TBD}
\providecommand{\GzEightClrARrtPlan}{\TBD}
\providecommand{\GzEightClrARrtExec}{\TBD}
\providecommand{\GzEightClrARrtTraj}{\TBD}
\providecommand{\GzEightClrARrtCFGT}{\TBD}
\providecommand{\GzEightClrARrtClear}{\TBD}
\providecommand{\GzEightClrARrtLink}{\TBD}
\providecommand{\GzEightClrBNpfN}{\TBD}
\providecommand{\GzEightClrBNpfUnreach}{\TBD}
\providecommand{\GzEightClrBNpfNoPath}{\TBD}
\providecommand{\GzEightClrBNpfSR}{\TBD}
\providecommand{\GzEightClrBNpfPlan}{\TBD}
\providecommand{\GzEightClrBNpfExec}{\TBD}
\providecommand{\GzEightClrBNpfTraj}{\TBD}
\providecommand{\GzEightClrBNpfCFGT}{\TBD}
\providecommand{\GzEightClrBNpfClear}{\TBD}
\providecommand{\GzEightClrBNpfLink}{\TBD}
\providecommand{\GzEightClrBRrtN}{\TBD}
\providecommand{\GzEightClrBRrtUnreach}{\TBD}
\providecommand{\GzEightClrBRrtNoPath}{\TBD}
\providecommand{\GzEightClrBRrtSR}{\TBD}
\providecommand{\GzEightClrBRrtPlan}{\TBD}
\providecommand{\GzEightClrBRrtExec}{\TBD}
\providecommand{\GzEightClrBRrtTraj}{\TBD}
\providecommand{\GzEightClrBRrtCFGT}{\TBD}
\providecommand{\GzEightClrBRrtClear}{\TBD}
\providecommand{\GzEightClrBRrtLink}{\TBD}
\providecommand{\AblEightMagGoalErr}{\TBD}
\providecommand{\AblEightMagSR}{\TBD}
\providecommand{\AblEightMagStall}{\TBD}
\providecommand{\AblEightMagCFGT}{\TBD}
\providecommand{\AblEightMagClearGT}{\TBD}
\providecommand{\AblEightMagGQ}{\TBD}
\providecommand{\AblEightMagBlkCFGT}{\TBD}
\providecommand{\GzEightClrCNpfN}{\TBD}
\providecommand{\GzEightClrCNpfUnreach}{\TBD}
\providecommand{\GzEightClrCNpfNoPath}{\TBD}
\providecommand{\GzEightClrCNpfSR}{\TBD}
\providecommand{\GzEightClrCNpfPlan}{\TBD}
\providecommand{\GzEightClrCNpfExec}{\TBD}
\providecommand{\GzEightClrCNpfTraj}{\TBD}
\providecommand{\GzEightClrCNpfCFGT}{\TBD}
\providecommand{\GzEightClrCNpfClear}{\TBD}
\providecommand{\GzEightClrCNpfLink}{\TBD}
\providecommand{\GzEightClrCRrtN}{\TBD}
\providecommand{\GzEightClrCRrtUnreach}{\TBD}
\providecommand{\GzEightClrCRrtNoPath}{\TBD}
\providecommand{\GzEightClrCRrtSR}{\TBD}
\providecommand{\GzEightClrCRrtPlan}{\TBD}
\providecommand{\GzEightClrCRrtExec}{\TBD}
\providecommand{\GzEightClrCRrtTraj}{\TBD}
\providecommand{\GzEightClrCRrtCFGT}{\TBD}
\providecommand{\GzEightClrCRrtClear}{\TBD}
\providecommand{\GzEightClrCRrtLink}{\TBD}
\providecommand{\GzEightClrDNpfN}{\TBD}
\providecommand{\GzEightClrDNpfUnreach}{\TBD}
\providecommand{\GzEightClrDNpfNoPath}{\TBD}
\providecommand{\GzEightClrDNpfSR}{\TBD}
\providecommand{\GzEightClrDNpfPlan}{\TBD}
\providecommand{\GzEightClrDNpfExec}{\TBD}
\providecommand{\GzEightClrDNpfTraj}{\TBD}
\providecommand{\GzEightClrDNpfCFGT}{\TBD}
\providecommand{\GzEightClrDNpfClear}{\TBD}
\providecommand{\GzEightClrDNpfLink}{\TBD}
\providecommand{\GzEightClrDRrtN}{\TBD}
\providecommand{\GzEightClrDRrtUnreach}{\TBD}
\providecommand{\GzEightClrDRrtNoPath}{\TBD}
\providecommand{\GzEightClrDRrtSR}{\TBD}
\providecommand{\GzEightClrDRrtPlan}{\TBD}
\providecommand{\GzEightClrDRrtExec}{\TBD}
\providecommand{\GzEightClrDRrtTraj}{\TBD}
\providecommand{\GzEightClrDRrtCFGT}{\TBD}
\providecommand{\GzEightClrDRrtClear}{\TBD}
\providecommand{\GzEightClrDRrtLink}{\TBD}
\providecommand{\AblEightDualGoalErr}{\TBD}
\providecommand{\AblEightDualSR}{\TBD}
\providecommand{\AblEightDualStall}{\TBD}
\providecommand{\AblEightDualCFGT}{\TBD}
\providecommand{\AblEightDualClearGT}{\TBD}
\providecommand{\AblEightDualGQ}{\TBD}
\providecommand{\AblEightDualBlkCFGT}{\TBD}
\providecommand{\GzEightDualNpfN}{\TBD}
\providecommand{\GzEightDualNpfUnreach}{\TBD}
\providecommand{\GzEightDualNpfNoPath}{\TBD}
\providecommand{\GzEightDualNpfSR}{\TBD}
\providecommand{\GzEightDualNpfPlan}{\TBD}
\providecommand{\GzEightDualNpfExec}{\TBD}
\providecommand{\GzEightDualNpfTraj}{\TBD}
\providecommand{\GzEightDualNpfCFGT}{\TBD}
\providecommand{\GzEightDualNpfClear}{\TBD}
\providecommand{\GzEightDualNpfLink}{\TBD}
\providecommand{\GzEightDualRrtN}{\TBD}
\providecommand{\GzEightDualRrtUnreach}{\TBD}
\providecommand{\GzEightDualRrtNoPath}{\TBD}
\providecommand{\GzEightDualRrtSR}{\TBD}
\providecommand{\GzEightDualRrtPlan}{\TBD}
\providecommand{\GzEightDualRrtExec}{\TBD}
\providecommand{\GzEightDualRrtTraj}{\TBD}
\providecommand{\GzEightDualRrtCFGT}{\TBD}
\providecommand{\GzEightDualRrtClear}{\TBD}
\providecommand{\GzEightDualRrtLink}{\TBD}
\providecommand{\AblEightDtGoalErr}{\TBD}
\providecommand{\AblEightDtSR}{\TBD}
\providecommand{\AblEightDtStall}{\TBD}
\providecommand{\AblEightDtCFGT}{\TBD}
\providecommand{\AblEightDtClearGT}{\TBD}
\providecommand{\AblEightDtGQ}{\TBD}
\providecommand{\AblEightDtBlkCFGT}{\TBD}
\providecommand{\GzEightDtNpfN}{\TBD}
\providecommand{\GzEightDtNpfUnreach}{\TBD}
\providecommand{\GzEightDtNpfNoPath}{\TBD}
\providecommand{\GzEightDtNpfSR}{\TBD}
\providecommand{\GzEightDtNpfPlan}{\TBD}
\providecommand{\GzEightDtNpfExec}{\TBD}
\providecommand{\GzEightDtNpfTraj}{\TBD}
\providecommand{\GzEightDtNpfCFGT}{\TBD}
\providecommand{\GzEightDtNpfClear}{\TBD}
\providecommand{\GzEightDtNpfLink}{\TBD}
\providecommand{\GzEightDtRrtN}{\TBD}
\providecommand{\GzEightDtRrtUnreach}{\TBD}
\providecommand{\GzEightDtRrtNoPath}{\TBD}
\providecommand{\GzEightDtRrtSR}{\TBD}
\providecommand{\GzEightDtRrtPlan}{\TBD}
\providecommand{\GzEightDtRrtExec}{\TBD}
\providecommand{\GzEightDtRrtTraj}{\TBD}
\providecommand{\GzEightDtRrtCFGT}{\TBD}
\providecommand{\GzEightDtRrtClear}{\TBD}
\providecommand{\GzEightDtRrtLink}{\TBD}
% ---- auto-filled by fill_results_macros.py (do not edit below) ----
\renewcommand{\ThreeNpfSR}{100.0}
\renewcommand{\ThreeRrtSR}{92.0}
\renewcommand{\ThreeNpfCF}{15.0}
\renewcommand{\ThreeRrtCF}{95.0}
\renewcommand{\ThreeNpfStall}{0.0}
\renewcommand{\ThreeNpfPlan}{155$\pm$21}
\renewcommand{\ThreeRrtPlan}{962$\pm$1119}
\renewcommand{\ThreeNpfTraj}{0.44$\pm$0.20}
\renewcommand{\ThreeRrtTraj}{0.83$\pm$0.40}
\renewcommand{\ThreeNpfClear}{0.008$\pm$0.003}
\renewcommand{\ThreeRrtClear}{0.007$\pm$0.002}
\renewcommand{\ThreeNpfSmooth}{0.10$\pm$0.07}
\renewcommand{\ThreeRrtSmooth}{0.73$\pm$0.06}
\renewcommand{\ThreeNpfCFGT}{95.0}
\renewcommand{\ThreeRrtCFGT}{85.0}
\renewcommand{\ThreeNpfClearGT}{0.087$\pm$0.054}
\renewcommand{\ThreeRrtClearGT}{0.086$\pm$0.039}
\renewcommand{\NumStarts}{20}
\renewcommand{\NumSeeds}{5}
\renewcommand{\TolGoal}{0.05}
\renewcommand{\EightNpfSR}{100.0}
\renewcommand{\EightRrtSR}{86.0}
\renewcommand{\EightNpfCF}{25.0}
\renewcommand{\EightRrtCF}{95.0}
\renewcommand{\EightNpfStall}{0.0}
\renewcommand{\EightNpfPlan}{124$\pm$32}
\renewcommand{\EightRrtPlan}{1036$\pm$1185}
\renewcommand{\EightNpfTraj}{0.45$\pm$0.24}
\renewcommand{\EightRrtTraj}{0.95$\pm$0.51}
\renewcommand{\EightNpfClear}{0.019$\pm$0.015}
\renewcommand{\EightRrtClear}{0.011$\pm$0.007}
\renewcommand{\EightNpfSmooth}{0.14$\pm$0.12}
\renewcommand{\EightRrtSmooth}{0.73$\pm$0.08}
\renewcommand{\EightNpfCFGT}{60.0}
\renewcommand{\EightRrtCFGT}{55.0}
\renewcommand{\EightNpfClearGT}{0.037$\pm$0.038}
\renewcommand{\EightRrtClearGT}{0.031$\pm$0.028}
\renewcommand{\AblFullSR}{100.0}
\renewcommand{\AblFullStall}{0.0}
\renewcommand{\ViewsSixtyfourSR}{100.0}
\renewcommand{\SegCleanSR}{100.0}
\renewcommand{\AblDecaySR}{100.0}
\renewcommand{\AblDecayStall}{0.0}
\renewcommand{\AblDecayGoalErr}{1.6}
\renewcommand{\AblDecayCFGT}{95.0}
\renewcommand{\AblDecayClearGT}{0.087}
\renewcommand{\AblDecayBlkCFGT}{40.0}
\renewcommand{\AblDecayGQ}{0.857}
\renewcommand{\AblNoAlignSR}{0.0}
\renewcommand{\AblNoAlignStall}{100.0}
\renewcommand{\AblNoAlignGoalErr}{1.4}
\renewcommand{\AblNoAlignCFGT}{0.0}
\renewcommand{\AblNoAlignClearGT}{--}
\renewcommand{\AblNoAlignBlkCFGT}{0.0}
\renewcommand{\AblNoAlignGQ}{0.845}
\renewcommand{\AblNoAnisoSR}{100.0}
\renewcommand{\AblNoAnisoStall}{0.0}
\renewcommand{\AblNoAnisoGoalErr}{0.7}
\renewcommand{\AblNoAnisoCFGT}{100.0}
\renewcommand{\AblNoAnisoClearGT}{0.094}
\renewcommand{\AblNoAnisoBlkCFGT}{60.0}
\renewcommand{\AblNoAnisoGQ}{0.760}
\renewcommand{\AblNoCSR}{100.0}
\renewcommand{\AblNoCStall}{0.0}
\renewcommand{\AblNoCGoalErr}{1.2}
\renewcommand{\AblNoCCFGT}{95.0}
\renewcommand{\AblNoCClearGT}{0.090}
\renewcommand{\AblNoCBlkCFGT}{20.0}
\renewcommand{\AblNoCGQ}{0.863}
\renewcommand{\AblNoGSR}{100.0}
\renewcommand{\AblNoGStall}{0.0}
\renewcommand{\AblNoGGoalErr}{1.0}
\renewcommand{\AblNoGCFGT}{95.0}
\renewcommand{\AblNoGClearGT}{0.092}
\renewcommand{\AblNoGBlkCFGT}{10.0}
\renewcommand{\AblNoGGQ}{0.694}
\renewcommand{\AblNoMagSR}{100.0}
\renewcommand{\AblNoMagStall}{0.0}
\renewcommand{\AblNoMagGoalErr}{1.2}
\renewcommand{\AblNoMagCFGT}{95.0}
\renewcommand{\AblNoMagClearGT}{0.091}
\renewcommand{\AblNoMagBlkCFGT}{5.0}
\renewcommand{\AblNoMagGQ}{0.853}
\renewcommand{\AblNoMvSR}{0.0}
\renewcommand{\AblNoMvStall}{100.0}
\renewcommand{\AblNoMvGoalErr}{62.8}
\renewcommand{\AblNoMvCFGT}{0.0}
\renewcommand{\AblNoMvClearGT}{--}
\renewcommand{\AblNoMvBlkCFGT}{0.0}
\renewcommand{\AblNoMvGQ}{0.000}
\renewcommand{\AblNoPotSR}{100.0}
\renewcommand{\AblNoPotStall}{0.0}
\renewcommand{\AblNoPotGoalErr}{1.1}
\renewcommand{\AblNoPotCFGT}{90.0}
\renewcommand{\AblNoPotClearGT}{0.088}
\renewcommand{\AblNoPotBlkCFGT}{10.0}
\renewcommand{\AblNoPotGQ}{0.852}
\renewcommand{\AblRadiusSR}{55.0}
\renewcommand{\AblRadiusStall}{45.0}
\renewcommand{\AblRadiusGoalErr}{0.9}
\renewcommand{\AblRadiusCFGT}{55.0}
\renewcommand{\AblRadiusClearGT}{0.123}
\renewcommand{\AblRadiusBlkCFGT}{25.0}
\renewcommand{\AblRadiusGQ}{0.848}
\renewcommand{\SegDilateSR}{100.0}
\renewcommand{\SegDilateStall}{0.0}
\renewcommand{\SegDilateGoalErr}{0.8}
\renewcommand{\SegDilateCFGT}{100.0}
\renewcommand{\SegDilateClearGT}{0.091}
\renewcommand{\SegDilateBlkCFGT}{30.0}
\renewcommand{\SegDilateGQ}{0.852}
\renewcommand{\SegDropSR}{100.0}
\renewcommand{\SegDropStall}{0.0}
\renewcommand{\SegDropGoalErr}{0.9}
\renewcommand{\SegDropCFGT}{100.0}
\renewcommand{\SegDropClearGT}{0.092}
\renewcommand{\SegDropBlkCFGT}{30.0}
\renewcommand{\SegDropGQ}{0.855}
\renewcommand{\SegErodeSR}{100.0}
\renewcommand{\SegErodeStall}{0.0}
\renewcommand{\SegErodeGoalErr}{0.9}
\renewcommand{\SegErodeCFGT}{95.0}
\renewcommand{\SegErodeClearGT}{0.091}
\renewcommand{\SegErodeBlkCFGT}{55.0}
\renewcommand{\SegErodeGQ}{0.859}
\renewcommand{\ViewsSixteenEpSR}{100.0}
\renewcommand{\ViewsSixteenEpStall}{0.0}
\renewcommand{\ViewsSixteenEpGoalErr}{2.2}
\renewcommand{\ViewsSixteenEpCFGT}{95.0}
\renewcommand{\ViewsSixteenEpClearGT}{0.088}
\renewcommand{\ViewsSixteenEpBlkCFGT}{35.0}
\renewcommand{\ViewsSixteenEpGQ}{0.820}
\renewcommand{\ViewsSixteenSR}{100.0}
\renewcommand{\ViewsSixteenStall}{0.0}
\renewcommand{\ViewsSixteenGoalErr}{0.9}
\renewcommand{\ViewsSixteenCFGT}{100.0}
\renewcommand{\ViewsSixteenClearGT}{0.091}
\renewcommand{\ViewsSixteenBlkCFGT}{35.0}
\renewcommand{\ViewsSixteenGQ}{0.833}
\renewcommand{\ViewsThirtytwoEpSR}{100.0}
\renewcommand{\ViewsThirtytwoEpStall}{0.0}
\renewcommand{\ViewsThirtytwoEpGoalErr}{2.5}
\renewcommand{\ViewsThirtytwoEpCFGT}{95.0}
\renewcommand{\ViewsThirtytwoEpClearGT}{0.089}
\renewcommand{\ViewsThirtytwoEpBlkCFGT}{35.0}
\renewcommand{\ViewsThirtytwoEpGQ}{0.853}
\renewcommand{\ViewsThirtytwoSR}{100.0}
\renewcommand{\ViewsThirtytwoStall}{0.0}
\renewcommand{\ViewsThirtytwoGoalErr}{1.7}
\renewcommand{\ViewsThirtytwoCFGT}{100.0}
\renewcommand{\ViewsThirtytwoClearGT}{0.089}
\renewcommand{\ViewsThirtytwoBlkCFGT}{30.0}
\renewcommand{\ViewsThirtytwoGQ}{0.848}
\renewcommand{\ViewsEightEpSR}{95.0}
\renewcommand{\ViewsEightEpStall}{5.0}
\renewcommand{\ViewsEightEpGoalErr}{1.0}
\renewcommand{\ViewsEightEpCFGT}{95.0}
\renewcommand{\ViewsEightEpClearGT}{0.104}
\renewcommand{\ViewsEightEpBlkCFGT}{45.0}
\renewcommand{\ViewsEightEpGQ}{0.754}
\renewcommand{\ViewsEightSR}{100.0}
\renewcommand{\ViewsEightStall}{0.0}
\renewcommand{\ViewsEightGoalErr}{0.3}
\renewcommand{\ViewsEightCFGT}{100.0}
\renewcommand{\ViewsEightClearGT}{0.096}
\renewcommand{\ViewsEightBlkCFGT}{30.0}
\renewcommand{\ViewsEightGQ}{0.788}
\renewcommand{\AblFullCFGT}{95.0}
\renewcommand{\AblFullClearGT}{0.087}
\renewcommand{\AblFullGQ}{0.859}
\renewcommand{\AblFullGoalErr}{1.6}
\renewcommand{\AblFullBlkCFGT}{25.0}
\renewcommand{\BlkThreeNpfSR}{100.0}
\renewcommand{\BlkThreeNpfCFGT}{25.0}
\renewcommand{\BlkThreeNpfClearGT}{0.010$\pm$0.008}
\renewcommand{\BlkThreeNpfTraj}{0.51$\pm$0.17}
\renewcommand{\BlkThreeRrtSR}{98.0}
\renewcommand{\BlkThreeRrtCFGT}{65.0}
\renewcommand{\BlkThreeRrtClearGT}{0.021$\pm$0.013}
\renewcommand{\BlkThreeRrtTraj}{0.82$\pm$0.25}
\renewcommand{\BlkEightNpfSR}{100.0}
\renewcommand{\BlkEightNpfCFGT}{25.0}
\renewcommand{\BlkEightNpfClearGT}{0.005$\pm$0.002}
\renewcommand{\BlkEightNpfTraj}{0.58$\pm$0.22}
\renewcommand{\BlkEightRrtSR}{100.0}
\renewcommand{\BlkEightRrtCFGT}{15.0}
\renewcommand{\BlkEightRrtClearGT}{0.008$\pm$0.006}
\renewcommand{\BlkEightRrtTraj}{0.91$\pm$0.32}
\renewcommand{\AblEightFullSR}{100.0}
\renewcommand{\AblEightFullStall}{0.0}
\renewcommand{\AblEightFullGoalErr}{2.4}
\renewcommand{\AblEightFullCFGT}{60.0}
\renewcommand{\AblEightFullClearGT}{0.037}
\renewcommand{\AblEightFullBlkCFGT}{25.0}
\renewcommand{\AblEightFullGQ}{0.755}
\renewcommand{\AblEightNoGSR}{100.0}
\renewcommand{\AblEightNoGStall}{0.0}
\renewcommand{\AblEightNoGGoalErr}{0.4}
\renewcommand{\AblEightNoGCFGT}{55.0}
\renewcommand{\AblEightNoGClearGT}{0.032}
\renewcommand{\AblEightNoGBlkCFGT}{10.0}
\renewcommand{\AblEightNoGGQ}{0.000}
\renewcommand{\AblEightNoMvSR}{0.0}
\renewcommand{\AblEightNoMvStall}{100.0}
\renewcommand{\AblEightNoMvGoalErr}{2.4}
\renewcommand{\AblEightNoMvCFGT}{0.0}
\renewcommand{\AblEightNoMvClearGT}{--}
\renewcommand{\AblEightNoMvBlkCFGT}{0.0}
\renewcommand{\AblEightNoMvGQ}{0.745}
\renewcommand{\AblEightNoAnisoSR}{100.0}
\renewcommand{\AblEightNoAnisoStall}{0.0}
\renewcommand{\AblEightNoAnisoGoalErr}{1.5}
\renewcommand{\AblEightNoAnisoCFGT}{60.0}
\renewcommand{\AblEightNoAnisoClearGT}{0.036}
\renewcommand{\AblEightNoAnisoBlkCFGT}{15.0}
\renewcommand{\AblEightNoAnisoGQ}{0.784}
\renewcommand{\AblEightDecaySR}{100.0}
\renewcommand{\AblEightDecayStall}{0.0}
\renewcommand{\AblEightDecayGoalErr}{2.4}
\renewcommand{\AblEightDecayCFGT}{60.0}
\renewcommand{\AblEightDecayClearGT}{0.037}
\renewcommand{\AblEightDecayBlkCFGT}{25.0}
\renewcommand{\AblEightDecayGQ}{0.755}
\renewcommand{\AblEightRadiusSR}{50.0}
\renewcommand{\AblEightRadiusStall}{50.0}
\renewcommand{\AblEightRadiusGoalErr}{0.8}
\renewcommand{\AblEightRadiusCFGT}{45.0}
\renewcommand{\AblEightRadiusClearGT}{0.053}
\renewcommand{\AblEightRadiusBlkCFGT}{5.0}
\renewcommand{\AblEightRadiusGQ}{0.744}
\renewcommand{\AblEightViewsSR}{100.0}
\renewcommand{\AblEightViewsStall}{0.0}
\renewcommand{\AblEightViewsGoalErr}{4.8}
\renewcommand{\AblEightViewsCFGT}{60.0}
\renewcommand{\AblEightViewsClearGT}{0.034}
\renewcommand{\AblEightViewsBlkCFGT}{5.0}
\renewcommand{\AblEightViewsGQ}{0.799}
\renewcommand{\AblEightClrSR}{100.0}
\renewcommand{\AblEightClrStall}{0.0}
\renewcommand{\AblEightClrGoalErr}{2.0}
\renewcommand{\AblEightClrCFGT}{55.0}
\renewcommand{\AblEightClrClearGT}{0.036}
\renewcommand{\AblEightClrBlkCFGT}{5.0}
\renewcommand{\AblEightClrGQ}{0.780}
\renewcommand{\AblEightMagSR}{100.0}
\renewcommand{\AblEightMagStall}{0.0}
\renewcommand{\AblEightMagGoalErr}{0.6}
\renewcommand{\AblEightMagCFGT}{65.0}
\renewcommand{\AblEightMagClearGT}{0.033}
\renewcommand{\AblEightMagBlkCFGT}{0.0}
\renewcommand{\AblEightMagGQ}{0.820}
\renewcommand{\AblEightDualSR}{100.0}
\renewcommand{\AblEightDualStall}{0.0}
\renewcommand{\AblEightDualGoalErr}{0.8}
\renewcommand{\AblEightDualCFGT}{60.0}
\renewcommand{\AblEightDualClearGT}{0.034}
\renewcommand{\AblEightDualBlkCFGT}{10.0}
\renewcommand{\AblEightDualGQ}{0.792}
\renewcommand{\AblEightDtSR}{100.0}
\renewcommand{\AblEightDtStall}{0.0}
\renewcommand{\AblEightDtGoalErr}{1.1}
\renewcommand{\AblEightDtCFGT}{55.0}
\renewcommand{\AblEightDtClearGT}{0.033}
\renewcommand{\AblEightDtBlkCFGT}{15.0}
\renewcommand{\AblEightDtGQ}{0.336}
\renewcommand{\GzThreeNpfN}{10}
\renewcommand{\GzThreeNpfUnreach}{1}
\renewcommand{\GzThreeNpfNoPath}{0}
\renewcommand{\GzThreeNpfSR}{100.0}
\renewcommand{\GzThreeNpfPlan}{0.29$\pm$0.05}
\renewcommand{\GzThreeNpfExec}{7.8$\pm$0.8}
\renewcommand{\GzThreeNpfTraj}{0.24$\pm$0.03}
\renewcommand{\GzThreeNpfCFGT}{100.0}
\renewcommand{\GzThreeNpfClear}{0.121}
\renewcommand{\GzThreeNpfLink}{0/91}
\renewcommand{\GzThreeRrtN}{10}
\renewcommand{\GzThreeRrtUnreach}{1}
\renewcommand{\GzThreeRrtNoPath}{1}
\renewcommand{\GzThreeRrtSR}{88.9}
\renewcommand{\GzThreeRrtPlan}{84.85$\pm$3.72}
\renewcommand{\GzThreeRrtExec}{1.8$\pm$0.3}
\renewcommand{\GzThreeRrtTraj}{0.25$\pm$0.03}
\renewcommand{\GzThreeRrtCFGT}{100.0}
\renewcommand{\GzThreeRrtClear}{0.121}
\renewcommand{\GzThreeRrtLink}{--}
\renewcommand{\GzEightNpfN}{20}
\renewcommand{\GzEightNpfUnreach}{3}
\renewcommand{\GzEightNpfNoPath}{0}
\renewcommand{\GzEightNpfSR}{88.2}
\renewcommand{\GzEightNpfPlan}{0.32$\pm$0.13}
\renewcommand{\GzEightNpfExec}{15.4$\pm$6.0}
\renewcommand{\GzEightNpfTraj}{0.55$\pm$0.20}
\renewcommand{\GzEightNpfCFGT}{26.7}
\renewcommand{\GzEightNpfClear}{0.008}
\renewcommand{\GzEightNpfLink}{0/2472}
\renewcommand{\GzEightRrtN}{20}
\renewcommand{\GzEightRrtUnreach}{1}
\renewcommand{\GzEightRrtNoPath}{14}
\renewcommand{\GzEightRrtSR}{26.3}
\renewcommand{\GzEightRrtPlan}{33.75$\pm$2.76}
\renewcommand{\GzEightRrtExec}{6.4$\pm$2.2}
\renewcommand{\GzEightRrtTraj}{0.79$\pm$0.27}
\renewcommand{\GzEightRrtCFGT}{20.0}
\renewcommand{\GzEightRrtClear}{0.004}
\renewcommand{\GzEightRrtLink}{0/3934}
\renewcommand{\GzEightRrtRunNpfN}{0}
\renewcommand{\GzEightRrtRunNpfUnreach}{0}
\renewcommand{\GzEightRrtRunNpfNoPath}{0}
\renewcommand{\GzEightRrtRunNpfSR}{0.0}
\renewcommand{\GzEightRrtRunNpfLink}{--}
\renewcommand{\GzEightRrtRunRrtN}{20}
\renewcommand{\GzEightRrtRunRrtUnreach}{1}
\renewcommand{\GzEightRrtRunRrtNoPath}{15}
\renewcommand{\GzEightRrtRunRrtSR}{21.1}
\renewcommand{\GzEightRrtRunRrtPlan}{29.61$\pm$6.83}
\renewcommand{\GzEightRrtRunRrtExec}{4.9$\pm$2.7}
\renewcommand{\GzEightRrtRunRrtTraj}{0.69$\pm$0.41}
\renewcommand{\GzEightRrtRunRrtCFGT}{0.0}
\renewcommand{\GzEightRrtRunRrtClear}{0.003}
\renewcommand{\GzEightRrtRunRrtLink}{--}
\renewcommand{\GzEightGtNpfN}{0}
\renewcommand{\GzEightGtNpfUnreach}{0}
\renewcommand{\GzEightGtNpfNoPath}{0}
\renewcommand{\GzEightGtNpfSR}{0.0}
\renewcommand{\GzEightGtNpfLink}{--}
\renewcommand{\GzEightGtRrtN}{20}
\renewcommand{\GzEightGtRrtUnreach}{3}
\renewcommand{\GzEightGtRrtNoPath}{0}
\renewcommand{\GzEightGtRrtSR}{88.2}
\renewcommand{\GzEightGtRrtPlan}{131.27$\pm$24.17}
\renewcommand{\GzEightGtRrtExec}{6.7$\pm$2.2}
\renewcommand{\GzEightGtRrtTraj}{0.84$\pm$0.28}
\renewcommand{\GzEightGtRrtCFGT}{100.0}
\renewcommand{\GzEightGtRrtClear}{0.034}
\renewcommand{\GzEightGtRrtLink}{--}
\renewcommand{\GzEightClrBNpfN}{8}
\renewcommand{\GzEightClrBNpfUnreach}{0}
\renewcommand{\GzEightClrBNpfNoPath}{0}
\renewcommand{\GzEightClrBNpfSR}{100.0}
\renewcommand{\GzEightClrBNpfPlan}{0.36$\pm$0.17}
\renewcommand{\GzEightClrBNpfExec}{15.0$\pm$6.9}
\renewcommand{\GzEightClrBNpfTraj}{0.60$\pm$0.24}
\renewcommand{\GzEightClrBNpfCFGT}{25.0}
\renewcommand{\GzEightClrBNpfClear}{0.010}
\renewcommand{\GzEightClrBNpfLink}{--}
\renewcommand{\GzEightClrBRrtN}{0}
\renewcommand{\GzEightClrBRrtUnreach}{0}
\renewcommand{\GzEightClrBRrtNoPath}{0}
\renewcommand{\GzEightClrBRrtSR}{0.0}
\renewcommand{\GzEightClrBRrtLink}{--}
\renewcommand{\GzEightClrANpfN}{8}
\renewcommand{\GzEightClrANpfUnreach}{1}
\renewcommand{\GzEightClrANpfNoPath}{0}
\renewcommand{\GzEightClrANpfSR}{100.0}
\renewcommand{\GzEightClrANpfPlan}{0.30$\pm$0.18}
\renewcommand{\GzEightClrANpfExec}{13.9$\pm$7.1}
\renewcommand{\GzEightClrANpfTraj}{0.54$\pm$0.23}
\renewcommand{\GzEightClrANpfCFGT}{28.6}
\renewcommand{\GzEightClrANpfClear}{0.006}
\renewcommand{\GzEightClrANpfLink}{--}
\renewcommand{\GzEightClrARrtN}{0}
\renewcommand{\GzEightClrARrtUnreach}{0}
\renewcommand{\GzEightClrARrtNoPath}{0}
\renewcommand{\GzEightClrARrtSR}{0.0}
\renewcommand{\GzEightClrARrtLink}{--}
\renewcommand{\GzEightClrCNpfN}{8}
\renewcommand{\GzEightClrCNpfUnreach}{0}
\renewcommand{\GzEightClrCNpfNoPath}{0}
\renewcommand{\GzEightClrCNpfSR}{87.5}
\renewcommand{\GzEightClrCNpfPlan}{0.70$\pm$0.29}
\renewcommand{\GzEightClrCNpfExec}{40.4$\pm$16.3}
\renewcommand{\GzEightClrCNpfTraj}{0.59$\pm$0.26}
\renewcommand{\GzEightClrCNpfCFGT}{14.3}
\renewcommand{\GzEightClrCNpfClear}{0.003}
\renewcommand{\GzEightClrCNpfLink}{--}
\renewcommand{\GzEightClrCRrtN}{0}
\renewcommand{\GzEightClrCRrtUnreach}{0}
\renewcommand{\GzEightClrCRrtNoPath}{0}
\renewcommand{\GzEightClrCRrtSR}{0.0}
\renewcommand{\GzEightClrCRrtLink}{--}
\renewcommand{\GzEightDualNpfN}{8}
\renewcommand{\GzEightDualNpfUnreach}{0}
\renewcommand{\GzEightDualNpfNoPath}{0}
\renewcommand{\GzEightDualNpfSR}{62.5}
\renewcommand{\GzEightDualNpfPlan}{0.50$\pm$0.21}
\renewcommand{\GzEightDualNpfExec}{12.6$\pm$4.5}
\renewcommand{\GzEightDualNpfTraj}{0.46$\pm$0.18}
\renewcommand{\GzEightDualNpfCFGT}{40.0}
\renewcommand{\GzEightDualNpfClear}{0.011}
\renewcommand{\GzEightDualNpfLink}{--}
\renewcommand{\GzEightDualRrtN}{0}
\renewcommand{\GzEightDualRrtUnreach}{0}
\renewcommand{\GzEightDualRrtNoPath}{0}
\renewcommand{\GzEightDualRrtSR}{0.0}
\renewcommand{\GzEightDualRrtLink}{--}
\renewcommand{\GzEightDtNpfN}{8}
\renewcommand{\GzEightDtNpfUnreach}{1}
\renewcommand{\GzEightDtNpfNoPath}{0}
\renewcommand{\GzEightDtNpfSR}{57.1}
\renewcommand{\GzEightDtNpfPlan}{1.52$\pm$0.62}
\renewcommand{\GzEightDtNpfExec}{36.5$\pm$15.5}
\renewcommand{\GzEightDtNpfTraj}{0.66$\pm$0.30}
\renewcommand{\GzEightDtNpfCFGT}{0.0}
\renewcommand{\GzEightDtNpfClear}{0.004}
\renewcommand{\GzEightDtNpfLink}{--}
\renewcommand{\GzEightDtRrtN}{0}
\renewcommand{\GzEightDtRrtUnreach}{0}
\renewcommand{\GzEightDtRrtNoPath}{0}
\renewcommand{\GzEightDtRrtSR}{0.0}
\renewcommand{\GzEightDtRrtLink}{--}

\maketitle
\thispagestyle{empty}
\pagestyle{empty}

\begin{abstract}
Reaching a 6-DoF grasp pose in clutter requires a collision-free trajectory, conventionally
obtained by reconstructing the scene in 3D and planning inside that reconstruction, at the cost of
its accuracy and compute. Potential fields learned directly from images remove that dependency but
inherit the classical weakness of artificial potential fields: where attractive and repulsive
gradients cancel, the descent grazes the obstacle instead of going around it, and can stall
short of the goal. We present
an SE(3) neural potential field learned from posed RGB images and supervised with a \emph{navigation
function}, the geodesic distance to the grasp through free space recovered from those same images
during training, which removes both failures. On two tabletop scenes, from obstacle-blocked starts
executed on a UR10, the field converges within $3$\,cm of the grasp from every start and every path
it executes is collision-free against the ground-truth geometry, against $25\,\%$ and $0\,\%$ under
image supervision alone; mean clearance rises from under a centimeter to $8.6$--$8.8$\,cm and
arm-link contacts fall from $20.6$--$50.4\,\%$ to $2.7$--$5.5\,\%$ of executed configurations.
Executed grasp success is \PoseVFourNpfSR\,\% and \PoseBookNpfSR\,\% on the two scenes, the residual
failures being refusals of the Cartesian executor rather than of the field. Planning takes about
$2$\,s against $67$--$133$\,s for RRT* on a reconstruction of the same images, though under a common
offline harness the two are comparable: the deployed margin is the cost of collision-checking a
dense reconstruction, not planner complexity.
\end{abstract}

\section{INTRODUCTION}
Motion planning for manipulation in unstructured scenes relies on a 3D understanding of the workspace~\cite{xiang2024grasping,leebron2025b4p}, obtained from depth sensors~\cite{ali2019multi}, multi-view reconstruction~\cite{izadi2011kinectfusion,mildenhall2021nerf}, or implicitly by vision-to-action policies~\cite{levine2016end}. Depth is costly and incomplete~\cite{9875218}, reconstruction is noisy and slow~\cite{bi2025jasrnet}, and end-to-end policies are hard to interpret~\cite{wang2023measuring}.
We take a middle road: inspired by neural radiance fields~\cite{mildenhall2021nerf}, we train a network directly from posed images to represent a \emph{neural potential field} (NPF) over end-effector poses whose negative gradient is the plan. No explicit 3D model is built; the field only has to be accurate along the trajectories the gripper will follow, and it remains interpretable as a cost landscape.

The 3D field is supervised through a novel 3D-to-2D projection (Sec.~\ref{sec:gradient3D-2D}) that projects its gradient into each training view and compares it with 2D potential gradients derived from instance segmentation (Sec.~\ref{sec:gradient2D}). Such image-derived attraction/repulsion targets, however, inherit the classical weakness of artificial potential fields: summed attractive and repulsive gradients cancel near obstacles, so the descent grazes surfaces rather than detouring around them, and can stall at spurious minima. We therefore additionally supervise the field with a \emph{navigation function}  the geodesic distance to the grasp through the configuration-space free set recovered from the same images (Sec.~\ref{sec:navfn})  which has no minimum other than the goal, and distill it into the field with point-sampled losses.
Our contributions are: (i)~an SE(3) neural potential field learned from images, defined through three rigidly attached gripper points so that position and orientation are planned jointly; (ii)~a 3D-to-2D potential-gradient projection that trains it from segmentation-derived 2D fields; (iii)~navigation-function supervision built from a visual hull of the training views, which removes stalls and spurious minima by construction and needs no depth, mesh or CAD model; and (iv)~executed Gazebo trials on the cluttered eight-object scene: from \PoseVFourNpfN{} obstacle-blocked start poses the navigation-function NPF reaches the grasp pose on \PoseVFourNpfSR\,\% of them, with \PoseVFourNpfCFGT\,\% of its executed paths collision-free against the ground-truth objects at $8.8$\,cm mean clearance, and plans each query in \PoseVFourNpfPlan\,s; the same field trained from images alone is collision-free on \PoseVFourNpfImgCFGT\,\% of its paths, while RRT* needs \PoseVFourRrtPlan\,s per query on a reconstruction of those images (\PoseVFourGtPlan\,s on ground-truth geometry) and reaches \PoseVFourRrtSR\,\% (\PoseVFourGtSR\,\%) of the starts.
% Our contributions are: (i)~an SE(3) neural potential field learned from images, defined through three rigidly attached gripper points so that position and orientation are planned jointly; (ii)~a 3D-to-2D potential-gradient projection that trains it from segmentation-derived 2D fields; (iii)~navigation-function supervision built from a visual hull of the training views, which removes stalls and spurious minima by construction and needs no depth, mesh, or CAD model; and (iv)~executed Gazebo experiments against RRT* on reconstructed and ground-truth geometry, in which the navigation-function NPF is collision-free on every executed path of both scenes and plans in about 2\,s against 67-133\,s. \MAS{It is vague. What are both scenes? Which method is 2s which 133s? I suggest name one experiment and explain numbers explicitly for that experimetn.}

\section{Related Work}
% \textbf{Potential fields and navigation functions.}
% \MAS{It is good if you can add an explanation of what readers will see.}
This paper lies at the intersection of potential-field methods, neural scene representations, and grasp and motion planning. We briefly review these three lines of work and position our approach with respect to each.

\subsection{Potential Fields and Navigation Functions}
Artificial potential fields~\cite{khatib1986real} turn planning into gradient descent on a goal attractor plus obstacle repulsors and remain attractive for reactive control~\cite{wang2023pre}, but the summed field has spurious local minima and requires hand-tuned weights, usually addressed with perturbations or hybrid escape strategies~\cite{Hao2022}. Navigation functions~\cite{rimon1992exact} remove the minima by construction on a known free space. We learn the field from images and supervise it with a navigation function computed from image-derived geometry, so the local-minimum problem is addressed at the target rather than at run time.

% \textbf{Planning in neural scene representations.} 
\subsection{Planning in Neural Scene Representations} 
NeRFs have been used as planning substrates by treating rendered density as a collision cost~\cite{adamkiewicz2022vision}, as a Poisson point process yielding probabilistically safe voxels~\cite{chen2024catnips}, or as a source of reachable sets~\cite{thapliyal2026safe}; grasp-oriented variants add object-level reasoning~\cite{wang2022hierarchical,huang2025benchmarking} or language grounding~\cite{makarova2025diffusionrl,zheng2024gaussiangrasper,xiang2024grasping}. All keep an explicit geometric proxy between the field and the planner. We instead use the NeRF machinery only as a differentiable projection operator: the learned quantity is the SE(3) planning cost itself, and no density threshold, occupancy map, or mesh is extracted at planning time.

% \textbf{Grasp and motion planning.} 
\subsection{Grasp and Motion Planning}
Sampling-based planners such as RRT* are general but take tens of seconds on reconstructed geometry, optimization-based planners need accurate models and initialization, and learned policies answer in milliseconds but need task-specific data.
Recent work couples grasp selection with motion generation~\cite{xiang2024grasping,leebron2025b4p,vu2025online,huang2025spatialrobograspgeneralizedrobotic,liang2025whole} or predicts discrete grasps end-to-end~\cite{zhao2025robot}. Our method forgoes reconstruction and yields the entire SE(3) trajectory from a single learned field.

\section{Background}
\subsection{Neural Radiance Field (NeRF)}
NeRF~\cite{mildenhall2021nerf} represents a scene by a density $\sigma_\theta(\bfx)$ and a view-dependent color $\bfc_\theta(\bfx,\hat{\bfd})$. Along a ray $\bfr(t)=\bfo+t\hat{\bfd}$ sampled at $t_1<\dots<t_N$, the color is rendered by quadrature,
\begin{equation}
  \hat{C}(\bfr)\approx\sum_{i=1}^{N} T_i\,\alpha_i\,\bfc_i,
\label{eq:nerf_discrete}
\end{equation}
with transmittance $T_i=\exp(-\sum_{j<i}\sigma_j\delta_j)$, opacity $\alpha_i=1-e^{-\sigma_i\delta_i}$, and $\delta_i=t_{i+1}-t_i$. 
The networks are trained with the photometric loss $\mathcal{L}_c=\frac{1}{|\calR|}\sum_{\bfr\in\calR}\|\hat{C}(\bfr)-C(\bfr)\|_2^2$
% (Eq.~\ref{eq:color_loss})
over all rays $\calR$ of the training images.
\label{eq:color_loss}
% \MAS{This is wrong cross-refrence, I would remove Eq. III, as you don't refer to it anywhere.}

\subsection{SE(3) Potential Fields}
A pose is $(x,q)$ with $x\in\mathbb{R}^3$ and a unit quaternion $q\in S^3$.Since $q$ and $-q$ represent the same orientation, orientation distances use $d_q(q_1,q_2)=2\arccos(|q_1^\top q_2|)$ and poses are compared with $d_{SE(3)}=\|x-x^*\|+c\,d_q(q,q^*)$, $c=0.1$\,m/rad 
% (Eq.~\ref{eq:se3_dist_bg}).
\label{eq:se3_dist_bg}
% \MAS{Again wrong label, remove Eq.}
The potential network never consumes $q$ directly: the pose is expanded into three rigidly attached gripper points, the center and the two finger tips, using the fixed finger offsets $o_l,o_r$ supplied with the grasp,
\begin{equation}
p_c=x,\qquad p_l=x+R(q)\,o_l,\qquad p_r=x+R(q)\,o_r,
\label{eq:three_points}
\end{equation}
and a shared point-wise multilayer perceptron (MLP) $\psi_\theta$ with sinusoidal encoding gives the pose potential
\begin{equation}
\phi_\theta(x,q)\defeq\psi_\theta(p_c)+\psi_\theta(p_l)+\psi_\theta(p_r).
\label{eq:three_point_potential}
\end{equation}
Gradients with respect to $x$ and $q$ follow by automatic differentiation through the rigid coupling, so the field can act differently on each finger while preserving the gripper geometry.

\section{Method}
Given posed RGB images of a scene, the robot segments the goal object and the obstacles with an instance detector, converts each mask into a 2D potential (attractive for the goal, repulsive for obstacles; Sec.~\ref{sec:gradient2D}), and trains one SE(3) neural potential field with a NeRF-like architecture whose rendered gradient must match these 2D fields (Sec.~\ref{sec:gradient3D-2D}). The NeRF color loss regularizes geometry, three SE(3) goal losses establish the grasp pose as the minimum (Sec.~\ref{sec:se3_losses}), and a navigation function distilled from the visual hull of the same views removes spurious minima (Sec.~\ref{sec:navfn}). Although the color loss is retained, no reconstruction is extracted: planning follows the learned field directly (Sec.~\ref{sec:planning}).

\begin{figure*}[!t]
    \centering
    \includegraphics[width=0.78\linewidth]{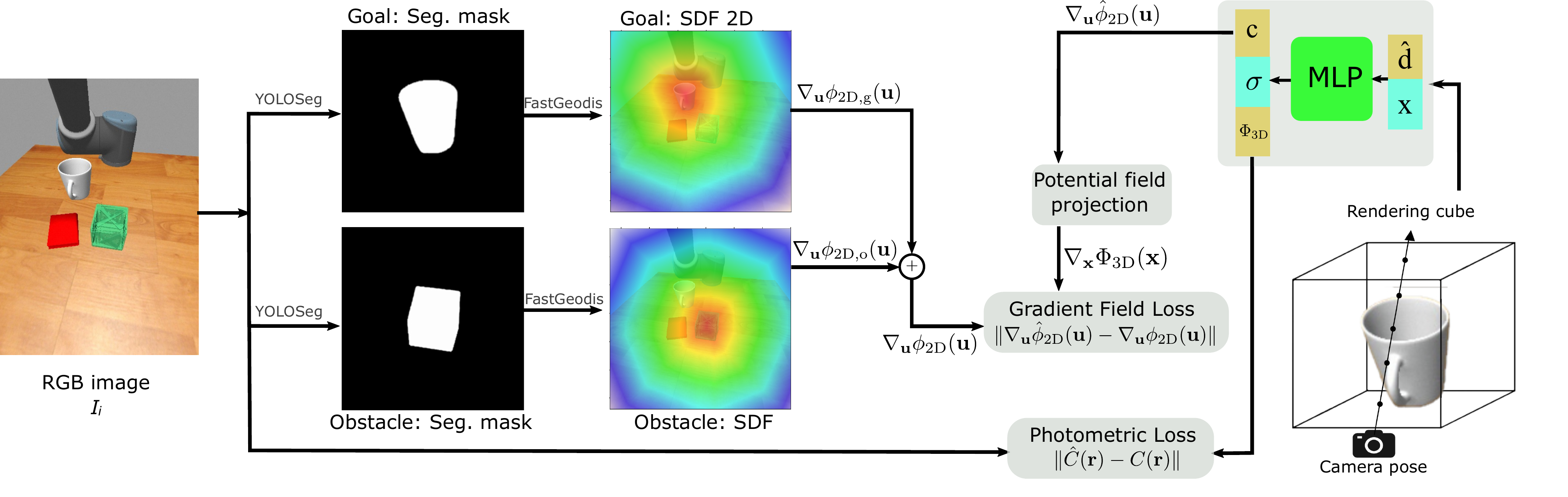}

        \caption{Training pipeline. Instance masks of each view become 2D signed-distance potentials (SDFs; attractive for the goal and repulsive for obstacles) whose gradients supervise the SE(3) potential network: a pose is expanded to the three gripper points, $\psi_\theta$ is evaluated at each, and the projected 3D gradients are aggregated along rays by density-weighted rendering. Color, projected-gradient, and SE(3) goal losses train the
field; the navigation-function losses of Sec.~\ref{sec:navfn} then refine it by point-sampled
distillation, without volume rendering.
% {\color{red}MAS:If you have actual image, 1-labels on arrows such as gradients are too small,  2-wrist image, have not been used in text, 3-put two losses close to eachother.}
}
\label{fig:overview}
\vspace{-3mm}
\end{figure*}

\subsection{Observed Potential Field in 2D}
\label{sec:gradient2D}
% For each image $I_i$ an instance detector yields binary masks $\{\bfM_o\}_{o=1}^O$, \MASE{where $o$ indexes the detected objects and one object ($g$) is the goal. 
% With $\mathbf{u}$ denoting a pixel coordinate and $d(\bfu,\bfM)$ the signed Euclidean distance transform of mask $M$} (negative inside), the goal field is attractive, $\phi_{\text{2D},g}(\bfu)=-\beta_g/(\max(d(\bfu,\bfM_g),0)+\epsilon)$, every other object is repulsive, $\phi_{\text{2D},o}(\bfu)=\alpha_r\exp(-\beta_O\,d(\bfu,\bfM_o))$, and the observed field is their sum $\phi_\text{2D}=\sum_o\phi_{\text{2D},o}$. The constants ($\alpha_r{=}100$, $\beta_g{=}2000$, $\epsilon{=}0.1$, decay $1/40$\,px$^{-1}$) are fixed once for all scenes.
% \MAS{1-is d(u,M) negative inside objects? If so, exp(-$\beta$d) become so large. 2-you defined $\beta_g$ what is $\beta_O$? }

For each image $I_i$, we segment the goal object and the obstacle objects using YOLOSeg, producing binary segmentation masks $\{\bfM_o\}_{o=1}^O = \text{YOLOSeg}(I_i)$ with $\bfM_o \in \{0,1\}^{H\times W}$ for all $o \in \{1, \dots, O\}$.
Among the segmented objects, one is the goal object, with ID $g$.
Using each segmentation  mask, we heuristically construct a 2D potential field for each object. The constants $\beta_g$, $\alpha_r$ and the repulsive decay are fixed once and shared across all scenes and objects (values in supplementary material); unlike classical APF deployments, no per-scene or per-object tuning is performed. Each per-object field has the resolution of the input image, $\phi_{\text{2D},o} \in \mathbb{R}^{H\times W}$, as do the combined field and its gradients.
For the goal object, we construct an attractive potential field that increases with the negative inverse of the signed distance field, $d(\bfu, \bfM_g)$,
\begin{align}
  \phi_\text{2D,g}(\bfu) &\defeq
\begin{cases}
  -\beta_g/(d(\bfu, \bfM_g) + \epsilon), & \text{if } d(\cdot) > 0 \\
-\beta_g/\epsilon, & \text{if } d(\cdot) \le 0.
\end{cases}
\end{align}%
where $g$ is the goal object id, and $\beta_g$ is a hyperparameter.
Here the signed distance field $d(\bfu, \bfM)$ is computed using the euclidean distance transform implemented in the \href{https://github.com/masadcv/FastGeodis}{FastGeodis} library~\cite{asad2022fastgeodis},
\begin{align}
  d(\bfu, \bfM) &\defeq \begin{cases}
    \min_{\bfv \mid \bfM[\bfv] = 1} \|\bfu - \bfv\|_2 &\text{ if } \bfM[\bfu] = 0\\
    -\min_{\bfv \mid \bfM[\bfv] = 0} \|\bfu - \bfv\|_2 &\text{ otherwise },
  \end{cases}
\end{align}
where the mask $\bfM[\bfv] = 1$ for all points $\bfv$ inside the object.
For all other obstacle objects, the repulsive potential field decays exponentially with the distance,
\begin{align}
  \phi_\text{2D,o}(\bfu) &\defeq
\alpha_r\exp\!\left(-\beta_O\, d(\bfu, \bfM_o)\right), & \forall\, o \neq g,
\end{align}%
where $\alpha_r$ and $\beta_O$ are hyperparameters.
Finally, the total observed 2D potential field is given by adding the potential fields of all objects, $\phi_\text{2D}(\bfu) \defeq \sum_{o=1}^O \phi_\text{2D,o}(\bfu)$.

\subsection{Potential Field Projection}
\label{sec:gradient3D-2D}
Each image only constrains the \emph{relative} attraction and repulsion of the end effector, so we relate the gradients of the 3D and 2D fields rather than their values. As in color rendering, the 3D gradient is integrated along rays with the opacity weights of Eq.~\eqref{eq:nerf_discrete}; the query orientation is fixed to the target $q^*$, so every sample $\bfx_i=\bfr(t_i)$ is expanded to the three gripper points, and
\begin{equation}
  \nabla_\bfu\hat{\phi}_\text{2D}(\bfu)\defeq\sum_{i=1}^N T_i\alpha_i\,w_{\text{aniso}}\!\!\sum_{k\in\{c,l,r\}}\!\!\proj_{2D}\!\bigl(\nabla_{p_k}\phi_{\theta,\text{3D}}\bigr),
\label{eq:grad_projection}
\end{equation}
where $\proj_{2D}(\bfy)=\pi(K\,{}^cR_w\,\bfy)$ projects a world-frame gradient through the intrinsics and rotation of the camera with $\pi(\bfv)=\bfv_{1:2}/\bfv_3$, and $w_{\text{aniso}}=\exp\bigl(\beta(1-(\hat{\bfd}^\top\bar{g}_i)^2)\bigr)$ emphasizes components perpendicular to the ray, where the image is informative. The projected-gradient loss is
\begin{equation}
\mathcal{L}_g=\frac{1}{|\calR|}\sum_{\bfr\in\calR}\bigl\|\nabla_\bfu\hat{\phi}_\text{2D}(\bfu)-\nabla_\bfu\phi_\text{2D}(\bfu)\bigr\|_2^2 .
\label{eq:gradient-field-loss}
\end{equation}
Supervising the projected gradient rather than the value makes the target invariant to the unobservable per-view offset of $\phi$.

\subsection{Target Grasp Pose}
\label{sec:target_pose}
The 6-DoF target $(x^*,q^*)$ and the finger offsets come from the AnyGrasp detector~\cite{fang2023anygrasp} on the training views and are stored as fixed, non-learnable buffers of the model, which removes the goal drift of the earlier position-only formulation and anchors the field to a physically meaningful grasp.

\subsection{SE(3) Goal-Aware Losses}
\label{sec:se3_losses}
Three losses make $\phi_{\theta,\text{3D}}$ a potential well centered at $(x^*,q^*)$. They sample poses from $\mathcal{B}$: positions from $\mathcal{N}(x^*,s_p^2I)$, $s_p=0.5$\,m, and orientations in a geodesic ball of radius $s_q=\pi$ around $q^*$ (20\,\% within $0.2$\,rad for local refinement). The rotation-vector gradient $\nabla_r\phi$ is the tangent-space projection of the quaternion gradient, and orientation is coupled to position with $s=0.2$\,m/rad.
\subsubsection*{Goal Potential Loss}
This loss pins the potential to zero at the target pose:
\begin{equation}
\mathcal{L}_{\text{pot}} = \lambda_{\text{pot}}\,
\phi_{\theta,\text{3D}}(x^*,q^*)^2,
\label{eq:grasp-pot}
\end{equation}
i.e. the three gripper points are at their target locations.

\subsubsection*{Gradient Magnitude Loss}
Near the target, the field must have a strong, non-vanishing slope.
We sample $N$ SE(3) poses $(x_i,q_i)$ around the target, convert each query quaternion to the relative rotation vector $r_i$, and penalize weak combined gradients:
\begin{equation}
\begin{split}
\mathcal{L}_{\text{mag}} = \lambda_{\text{mag}}\,
\mathbb{E}_{(x_i,q_i)\sim \mathcal{B}} \biggl[
\max\biggl(0,\, \gamma_{\min} \\
{}- \sqrt{\|\nabla_x \phi\|^2 + (s\|\nabla_r \phi\|)^2}
\biggr)
\biggr],
\end{split}
\label{eq:gradient_magnitude}
\end{equation}
where $\gamma_{\min}=0.1$, $s_p=0.5$~m, $s_q=\pi$~rad, and $\mathcal{B}$ is the sampler defined above, which mixes $20\%$ local orientation perturbations with $80\%$ samples up to the full geodesic radius.

\subsubsection*{Gradient Alignment Loss}
The negative gradient should point toward the target in the coupled position-rotation-vector space.
Let $g=[\nabla_x\phi^\top, s\nabla_r\phi^\top]^\top$
and $t=[(x^*-x)^\top, -s\,r^\top]^\top$, where $-r$ is the geodesic direction back to the target orientation.
Then:
\begin{equation}
\mathcal{L}_{\text{align}} = \lambda_{\text{align}}\,
\mathbb{E}\left[
1 - \frac{g^\top t}{\|g\|\|t\|}
\right].
\label{eq:gradient_alignment}
\end{equation}

Supervising the \emph{projected gradient} rather than the raw potential value makes the target invariant to the unknown per-view offset of $\phi$ (only differences of $\phi$ are observable in a projection); the loss nevertheless compares the full gradient vectors, so both the direction and the magnitude of the learned field are constrained, and the magnitude carries the distance-to-obstacle information.
Default weights are
$\lambda_{\text{pot}}=50$,
$\lambda_{\text{mag}}=10$, and
$\lambda_{\text{align}}=50$.

\subsubsection*{Orientation Bowl Loss}
Because $R(q)=R(-q)$, the three-point potential is double-cover invariant and needs no term against a minimum at $-q^*$; we nevertheless keep a hinge regularizer $\mathcal{L}_\text{bowl}$ that requires $\phi_{\theta,\text{3D}}(x,q)\ge\|x-x^*\|+d_q(q,q^*)$ on poses drawn over the full orientation ball, so that $\phi$ grows with distance from the target and large initial orientation errors receive a usable descent direction.
\label{eq:orientation_bowl}
A multi-view projection-consistency term on the goal center 
% (Eq.~\eqref{eq:multiview}, $\lambda_\text{mv}$)
is retained from the position-only model but disabled when the grasp buffer is used.
\label{eq:multiview}
% \MAS{No need for Eq. reference and label.}

\subsection{Network Training}
The network outputs $\sigma_\theta$, $\bfc_\theta$ and $\phi_{\theta,\text{3D}}$ and is trained on
\begin{equation}
\begin{split}
\mathcal{L}_{\text{total}}={}&\lambda_c\mathcal{L}_c+\lambda_g\mathcal{L}_g+\mathcal{L}_{\text{pot}}+\mathcal{L}_{\text{mag}}\\
&+\mathcal{L}_{\text{align}}+\lambda_{\text{bowl}}\mathcal{L}_{\text{bowl}}+\lambda_{\text{mv}}\mathcal{L}_{\text{mv}},
\end{split}
\label{eq:total_loss}
\end{equation}
with the SE(3) terms updating only the potential network; gradients are clipped to unit norm and accumulated over four mini-batches. The complete architecture and training configuration (4-layer, width-512 point MLP, Adam $5{\times}10^{-4}$, 1024 rays per step, 8 epochs, loss weights $\lambda_c{=}1,\lambda_g{=}2,\lambda_\text{pot}{=}50,\lambda_\text{mag}{=}10,\lambda_\text{align}{=}50,\lambda_\text{bowl}{=}20$) is given in the supplementary material; nothing is tuned per scene except the target-object class.

\subsection{Navigation-Function Supervision: Removing Stalls and Spurious Minima}
\label{sec:navfn}
A potential field assembled from a goal attractor and obstacle repulsors, the construction behind the
image-space targets above, inherits the classical weakness of artificial potential fields: wherever the
attractive and repulsive gradients cancel, the field has a \emph{spurious} minimum and gradient descent
stalls short of the goal, typically in the shadow of an obstacle or in a corridor between two of them.
In our executed trials, this appears as the Stall column of Table~\ref{tab:mesh_comparison_pose}, and the
same cancellation is what lets the descent graze surfaces on the approach: the field only pushes
\emph{away}, it never says \emph{around}.

A navigation function~\cite{rimon1992exact} removes this failure by construction: it is a scalar field
on the free space whose only minimum is the goal, so following its negative gradient from any free
configuration reaches the goal without entering an obstacle. We supervise the learned field with such a
function. Obstacle geometry is recovered from the same images used for training, as the visual hull of
the detector masks over the posed views (no depth, no reconstruction mesh); the hull is inflated by the
gripper radius (7\,cm) and the table becomes a floor, which yields the configuration-space free set
$\mathcal{F}$. The navigation function is the geodesic distance to the grasp pose inside $\mathcal{F}$,
$D(x)=\min_{\gamma\subset\mathcal{F}}\operatorname{len}(\gamma(x\!\to\!x^*))$, computed once by Dijkstra
on a 1\,cm voxel grid. This geometry is a \emph{training-time} construct only: the hull is a coarse
silhouette-carved occupancy envelope, built from the masks already used by the 2D supervision and never refined into a surface.
Neither the hull nor $D$ exists at run time; planning remains gradient descent on the learned field alone, with no geometric model in the loop. 
What the method avoids is a reconstruction on which the \emph{planner} depends, not the use of multi-view
consistency during training.
$D$ has no spurious minima: for every $x\neq x^*$ the first step of the shortest path lowers $D$, so a descent direction exists everywhere in $\mathcal{F}$, and level sets of $D$ bend
around obstacles instead of terminating on them. Clearance is not a tuned constant but a property of
$\mathcal{F}$: every path of $D$ keeps at least the inflation radius from the observed surfaces.

The field distills $D$ through point-sampled losses that need no volume rendering, so the supervision is
placed exactly where the obstacles are. At random poses $(x,q)$ in $\mathcal{F}$ we require
(i)~\emph{direction}: $1-\cos\!\big(\nabla_x\phi_{\theta,\text{3D}}(x,q),\,\nabla D(x)\big)$;
(ii)~\emph{descent progress}: $\max\!\big(0,\,1-\nabla_x\phi_{\theta,\text{3D}}\!\cdot\!\hat{\nabla}D\big)^2$,
a minimum slope along the navigation direction, which is the term that forbids flat regions and hence
stalls, and is sampled preferentially where the current slope is smallest; (iii)~a scale term that keeps
$\|\nabla_x\phi_{\theta,\text{3D}}\|$ at the value of the image-trained field; and (iv)~a goal-basin
anchor that preserves the learned grasp basin within 10\,cm of $x^*$. The repulsive head is trained to
the bounded distance potential $\rho(\mathrm{sdf}(x))$, which makes it a clearance predictor at run time.

\subsection{Trajectory Planning Using the SE(3) Neural Potential Field}
\label{sec:planning}
Planning is damped gradient descent on $\phi_{\theta,\text{3D}}$: position and quaternion gradients are obtained by automatic differentiation through the three-point mapping, smoothed by an exponential moving average and momentum, clipped, and applied with a step that shrinks within $0.2$\,m and $0.1$\,m of the target, the quaternion being re-normalized after every update. The rollout stops when $d_{SE(3)}<\epsilon$. At run time only this field is evaluated; the learned repulsive head of Sec.~\ref{sec:navfn} is not needed for steering.

\section{Experimental Evaluation}
\subsection{Datasets}
Experiments run in Gazebo with a 6-DoF UR10 and a Robotiq three-point gripper. Two earlier position-only scenes (\textbf{Balls}, \textbf{3 Obj.}) are retained in Table~\ref{tab:comparison} for reference. For the 6-DoF evaluation we collected two RGB-D scenes in which the end effector must reach a \emph{grasp pose}: \textbf{3 Obj.\ (book)}, where the book is the target and a cup and a crate are obstacles, and \textbf{8 Obj.\ (orange)}, a cluttered tabletop of eight everyday objects with an orange as the target (\PoseBookViews{} and \PoseVFourViews{} training views, camera orbit centered on the target). The free corridor around the target is wider than the gripper (nearest obstacle \PoseBookRing\,cm and \PoseVFourRing\,cm from the goal, $9$\,cm gripper). Real-robot rollouts use the setup of Fig.~\ref{fig:robot_exp}.

\subsection{Evaluation Metrics}
% \MAS{update this paragraph, to include information you need to cut from Tables 1 and 2. for instance "3,cm and \(15^\circ\)", "Link coll., the fraction of executed arm configurations with any link in collision with the ground-truth objects. ... ", "This is computed by forward-kinematics replay of the recorded joint trajectories, modeling each link as a capsule between consecutive joint origins and the gripper as a capsule along the tool axis."   }
We evaluate trajectory-planning performance using the following metrics~\cite{yang2023benchmarking,huang2025benchmarking}: 
success rate (SR), collision-free rate, trajectory length, planning time, execution time, final orientation error, and obstacle clearance.
A trial is successful if the final end-effector pose is within 3 cm and $15^\circ$ of the target grasp pose. 
The collision-free rate against ground-truth geometry (CF$_\text{GT}$) is the fraction of executed TCP trajectories that do not intersect any ground-truth obstacle mesh, excluding the target object and support table. 
Clearance is the minimum TCP-to-obstacle distance along the trajectory. 
To additionally evaluate whole-arm safety, Link coll. reports the fraction of executed arm configurations in which any robot link intersects a ground-truth obstacle. This is computed by forward-kinematics replay of the recorded joint trajectories, modeling each link as a capsule between consecutive joint origins and the gripper as a capsule along the tool axis. Trajectory length is the accumulated Euclidean distance between consecutive TCP waypoints; planning time measures the time required to generate a trajectory, and execution time measures the corresponding robot motion time. Orientation error (OE) is the geodesic quaternion distance $d_q(q_{\text{final}},q^*)$ between the final and target orientations.

% we report the success rate (final pose within $d_{SE(3)}$ of the target, $c=0.1$\,m/rad), trajectory length, planning and execution time, final orientation error, and the collision-free rate CF$_\text{GT}$: the fraction of \emph{executed} end-effector paths that never intersect the simulator's true object meshes, audited post hoc and independently of the geometry either planner used, which separates ``the planner believed it was safe'' from ``the robot was safe''.

\subsection{Baseline and Comparison Methods}
We compare our NPF (SE(3) field learned from multi-view images, trajectories by damped gradient descent, Sec.~IV) with the reconstruct-then-plan pipeline: a scene mesh from a vanilla NeRF trained on the same RGB-D views (any reconstruction method could be substituted~\cite{leroy2024grounding,wang2025vggt}), the target from YOLO~\cite{yolo} and AnyGrasp, and RRT*~\cite{noreen2016optimal} with a length-plus-clearance cost and a 2 cm collision margin on that mesh (target geometry excluded). Unlike our earlier evaluation, which gave the baseline the ground-truth models, the baseline here plans on the mesh the pipeline actually produced, and the ground-truth-geometry variant is reported separately as an upper bound; both planners therefore consume the same perception, which is the only comparison that isolates the planner.

\section{Results and Analysis}

\subsection{Quantitative Results}

\begin{table}[h!]
\caption{
% Executed in Gazebo (UR10, Robotiq three-point gripper, MoveIt Cartesian execution; mean$\pm$std across successful starts; RRT* rows of a scene come from the same session as its image-supervised NPF row). \PoseVFourNpfN{} SE(3) start poses per scene, goal = AnyGrasp grasp pose, success within 3\,cm and $15^\circ$. CF$_\text{GT}$: executed TCP path never intersects a ground-truth object mesh (goal object excluded); Link coll.: fraction of executed arm configurations with any link in collision with the
% ground-truth objects, from a forward-kinematics replay of the recorded joint trajectories in which
% each link is modeled as a capsule between consecutive joint origins and the gripper as a capsule
% along the tool axis (a conservative envelope of the true link geometry); OE: final orientation error. RRT* plans on the NeRF-reconstructed mesh with a 2\,cm margin or, in the ``GT geometry'' rows, on the true meshes.{\color{red}MAS: Too long. Keep some here, move some to main text.}
Executed trajectory-planning performance in Gazebo.
Comparison of the image-supervised NPF, navigation-function NPF, and RRT* using reconstructed or ground-truth geometry over 10 SE(3) start poses per scene. Success requires a final pose within 3 cm and \(15^\circ\) of the target. $N_\text{exec}$ : trials in which the arm executed a motion; in the remainder no motion took place
(the Cartesian executor aborted the plan, or, for the RRT* rows, the planner found no path); CF$_\text{GT}$ denotes collision-free executed TCP paths; Link coll. denotes the percentage of executed arm configurations in collision; OE is the final orientation error; $^\ddagger$RRT* interpolates to the goal quaternion, so its final orientation error is zero by construction rather than by search. Values are mean\(\pm\)std across successful starts.}

\label{tab:comparison}
\centering
\setlength{\tabcolsep}{3pt}
\resizebox{\columnwidth}{!}{%
\begin{tabular}{l l r r r r r r r r}
 \toprule
 Scene & Method & SR (\%) & $N_\text{exec}$ & CF$_\text{GT}$ (\%) & Plan.\ T.\ (s) & Exec.\ T.\ (s) & Traj.\ (m) & Link coll. & OE ($^\circ$)\\
 \midrule
 \midrule
 \midrule
 3 Obj.\ (book) & Recon.\ mesh + RRT* & \PoseBookRrtSR & \PoseBookRrtExecN/\PoseBookNpfN & \PoseBookRrtCFGT & \PoseBookRrtPlan & \PoseBookRrtExec & \PoseBookRrtTraj & \PoseBookRrtLink & $0.0^{\ddagger}$ \\
 3 Obj.\ (book) & GT geometry + RRT* & \PoseBookGtSR & \PoseBookGtExecN/\PoseBookNpfN & \PoseBookGtCFGT & \PoseBookGtPlan & \PoseBookGtExec & \PoseBookGtTraj & \PoseBookGtLink & $0.0^{\ddagger}$ \\
 3 Obj.\ (book) & NPF SE(3), image supervision only (prior) & \PoseBookNpfImgSR & \PoseBookNpfImgExecN/\PoseBookNpfN & \PoseBookNpfImgCFGT & \PoseBookNpfImgPlan & \PoseBookNpfImgExec & \PoseBookNpfImgTraj & \PoseBookNpfImgLink & \PoseBookNpfImgOE \\
 3 Obj.\ (book) & NPF SE(3) + nav.\ function (ours) & \PoseBookNpfSR & \PoseBookNpfExecN/\PoseBookNpfN & \PoseBookNpfCFGT & \PoseBookNpfPlan & \PoseBookNpfExec & \PoseBookNpfTraj & \PoseBookNpfLink & \PoseBookNpfOE \\
 \midrule
 8 Obj.\ (orange) & Recon.\ mesh + RRT* & \PoseVFourRrtSR & \PoseVFourRrtExecN/\PoseVFourNpfN & \PoseVFourRrtCFGT & \PoseVFourRrtPlan & \PoseVFourRrtExec & \PoseVFourRrtTraj & \PoseVFourRrtLink & $0.0^{\ddagger}$ \\
 8 Obj.\ (orange) & GT geometry + RRT* & \PoseVFourGtSR & \PoseVFourGtExecN/\PoseVFourNpfN & \PoseVFourGtCFGT & \PoseVFourGtPlan & \PoseVFourGtExec & \PoseVFourGtTraj & \PoseVFourGtLink & $0.0^{\ddagger}$ \\
 8 Obj.\ (orange) & NPF SE(3), image supervision only (prior) & \PoseVFourNpfImgSR & \PoseVFourNpfImgExecN/\PoseVFourNpfN & \PoseVFourNpfImgCFGT & \PoseVFourNpfImgPlan & \PoseVFourNpfImgExec & \PoseVFourNpfImgTraj & \PoseVFourNpfImgLink & \PoseVFourNpfImgOE \\
 8 Obj.\ (orange) & NPF SE(3) + nav.\ function (ours) & \PoseVFourNpfSR & \PoseVFourNpfExecN/\PoseVFourNpfN & \PoseVFourNpfCFGT & \PoseVFourNpfPlan & \PoseVFourNpfExec & \PoseVFourNpfTraj & \PoseVFourNpfLink & \PoseVFourNpfOE \\
 \bottomrule
\end{tabular}%
}
% \caption{Executed in Gazebo (UR10, Robotiq three-point gripper, MoveIt Cartesian execution; mean$\pm$std across successful starts; RRT* rows of a scene come from the same session as its image-supervised NPF row). \PoseVFourNpfN{} SE(3) start poses per scene, goal = AnyGrasp grasp pose, success within 3\,cm and $15^\circ$. CF$_\text{GT}$: executed TCP path never intersects a ground-truth object mesh (goal object excluded); Link coll.: fraction of executed arm configurations with any link in collision with the
% ground-truth objects, from a forward-kinematics replay of the recorded joint trajectories in which
% each link is modeled as a capsule between consecutive joint origins and the gripper as a capsule
% along the tool axis (a conservative envelope of the true link geometry); OE: final orientation error. RRT* plans on the NeRF-reconstructed mesh with a 2\,cm margin or, in the ``GT geometry'' rows, on the true meshes.}
% \label{tab:comparison}
\vspace{-3mm}
\end{table}

\begin{table*}[t]
\caption{
% Planning-only protocol: \emph{every} row is produced and scored in the same offline harness
% (single CPU, no simulator in the loop), on the same start sets ($N$ starts per scene; the RRT* rows
% average 5 random seeds per start), against the same geometry and with the same collision test a path
% is collision-free if no segment penetrates the reconstruction (CF-SR$_\text{mesh}$) or the ground-truth
% objects (CF-SR$_\text{GT}$, goal object and support table excluded); Clear.$_\text{GT}$ is the minimum
% TCP distance to those objects. Values are mean$\pm$std across starts. Planning times are therefore
% comparable across rows here but \emph{not} with Table~\ref{tab:comparison}, where planning runs inside
% the simulator against the dense mesh with gripper-body checks at every waypoint; execution times exist
% only in that deployed setting and are reported there. 
% {\color{red}MAS: Too long. Keep some here, move some to main text.}
\textbf{Planning-only comparison under a common offline protocol.} All methods are evaluated on the same start sets and collision geometry using a single-CPU harness, with RRT* averaged over five random seeds per start. CF-SR$_\text{mesh}$ and CF-SR$_\text{GT}$ denote collision-free success against reconstructed and ground-truth geometry, respectively; Clear.$_\text{GT}$ is the minimum TCP distance to ground-truth obstacles. Values are mean\(\pm\)std across starts.}

\label{tab:mesh_comparison_pose}
\centering
\setlength{\tabcolsep}{4pt}
\resizebox{\textwidth}{!}{%
\begin{tabular}{l l r r r r r r r}
 \toprule
 Scene & Method & $N$ & SR (\%) & CF-SR$_\text{mesh}$ (\%) & CF-SR$_\text{GT}$ (\%) & Stall (\%) & Traj.\ (m) & Clear.$_\text{GT}$ (m)\\
 \midrule
 3 Obj.\ (book) & Recon.\ mesh + RRT* (5 seeds) & \PoseBookRrtSwN & \PoseBookRrtSwSR & \PoseBookRrtSwCFmesh & \PoseBookRrtSwCFGT & \PoseBookRrtSwStall & \PoseBookRrtSwTraj & \PoseBookRrtSwClearGT \\
 3 Obj.\ (book) & NPF SE(3), image supervision only (prior) & \PoseBookNpfImgOffN & \PoseBookNpfImgOffSR & \PoseBookNpfImgOffCFmesh & \PoseBookNpfImgOffCFGT & \PoseBookNpfImgOffStall & \PoseBookNpfImgOffTraj & \PoseBookNpfImgOffClearGT \\
 3 Obj.\ (book) & NPF SE(3) + nav.\ function (ours) & \PoseBookNpfOffN & \PoseBookNpfOffSR & \PoseBookNpfOffCFmesh & \PoseBookNpfOffCFGT & \PoseBookNpfOffStall & \PoseBookNpfOffTraj & \PoseBookNpfOffClearGT \\
 \midrule
 8 Obj.\ (orange) & Recon.\ mesh + RRT* (5 seeds) & \PoseVFourRrtSwN & \PoseVFourRrtSwSR & \PoseVFourRrtSwCFmesh & \PoseVFourRrtSwCFGT & \PoseVFourRrtSwStall & \PoseVFourRrtSwTraj & \PoseVFourRrtSwClearGT \\
 8 Obj.\ (orange) & NPF SE(3), image supervision only (prior) & \PoseVFourNpfImgOffN & \PoseVFourNpfImgOffSR & \PoseVFourNpfImgOffCFmesh & \PoseVFourNpfImgOffCFGT & \PoseVFourNpfImgOffStall & \PoseVFourNpfImgOffTraj & \PoseVFourNpfImgOffClearGT \\
 8 Obj.\ (orange) & NPF SE(3) + nav.\ function (ours) & \PoseVFourNpfOffN & \PoseVFourNpfOffSR & \PoseVFourNpfOffCFmesh & \PoseVFourNpfOffCFGT & \PoseVFourNpfOffStall & \PoseVFourNpfOffTraj & \PoseVFourNpfOffClearGT \\
 \bottomrule
\end{tabular}}
% \caption{Planning-only protocol: \emph{every} row is produced and scored in the same offline harness
% (single CPU, no simulator in the loop), on the same start sets ($N$ starts per scene; the RRT* rows
% average 5 random seeds per start), against the same geometry and with the same collision test a path
% is collision-free if no segment penetrates the reconstruction (CF-SR$_\text{mesh}$) or the ground-truth
% objects (CF-SR$_\text{GT}$, goal object and support table excluded); Clear.$_\text{GT}$ is the minimum
% TCP distance to those objects. Values are mean$\pm$std across starts. Planning times are therefore
% comparable across rows here but \emph{not} with Table~\ref{tab:comparison}, where planning runs inside
% the simulator against the dense mesh with gripper-body checks at every waypoint; execution times exist
% only in that deployed setting and are reported there.}
% \label{tab:mesh_comparison_pose}
\vspace{-3mm}
\end{table*}
Tables~\ref{tab:comparison} and~\ref{tab:mesh_comparison_pose} compare, on the same start poses and the same perception, the image-supervised NPF, the NPF with navigation-function supervision, and RRT* planning on the reconstructed mesh (with the ground-truth-geometry variant as an upper bound). 
Table~\ref{tab:mesh_comparison_pose} repeats the comparison offline, without the simulator, with RRT* over five seeds per start.  
Image supervision alone, which extends the supervision of the earlier position-only formulation to SE(3), already reaches the grasp reliably and plans two orders of magnitude faster than RRT* (0.8-1.2\,s against 67-133\,s), but it does not produce a detour: on both scenes the descent drives toward the goal and passes within half a centimeter of the obstacle blocking the start, and the object it touches is precisely that blocker in 14 of 14 book and 14 of 20 eight-object rollouts, leaving its executed paths collision-free in only 25\,\% (8 objects) and 0\,\% (book) of trials. The same supervision is collision-free on 95\,\% of \emph{unblocked} starts but only 25\,\% of blocked ones in the position-only setting, so the grazing is a property of image-space repulsion a thin band at the projected mask boundary rather than of the SE(3) extension or the three-point gripper model; a fixed-start protocol hides it because the straight-line approach is already clear. The navigation function changes the behavior rather than the margin, and without changing the run-time procedure: every executed path in both scenes is collision-free against the ground-truth geometry, mean clearance rises to 8.6-8.8\,cm (against 2-3\,cm for RRT*, which plans with a 2\,cm margin), arm-body contacts fall from 20.6\,\% to 5.5\,\% of configurations, and planning stays at about 2\,s. 
The planning-only protocol of Table~\ref{tab:mesh_comparison_pose}
% , where every planner runs offline on one CPU against a lightweight point-sampled checker, 
isolates this safety gain from the run-time: there the separation is entirely in collision avoidance.6-2.6\,s), and the separation is entirely in collision avoidance, the navigation-function field reaches every start of both scenes without penetrating an object, at 8.2-9.7\,cm clearance, against 1.8-2.1\,cm for RRT* (which plans to a 2\,cm margin) and under 0.5\,cm for image supervision alone. The deployed speed-up reflects the cost of querying a dense reconstruction with gripper-body checks at every waypoint; the safety gain holds in both settings. 
Executed success is 90\,\% in the 8-object scene and is executor-bounded in the book scene (Sec.~\ref{sec:failures}). Execution is slower than RRT* (5\,s against 3\,s) because of conservative gradient steps, and paths are of comparable length.

\begin{figure*}[t]
\centering
\includegraphics[width=0.70\textwidth]{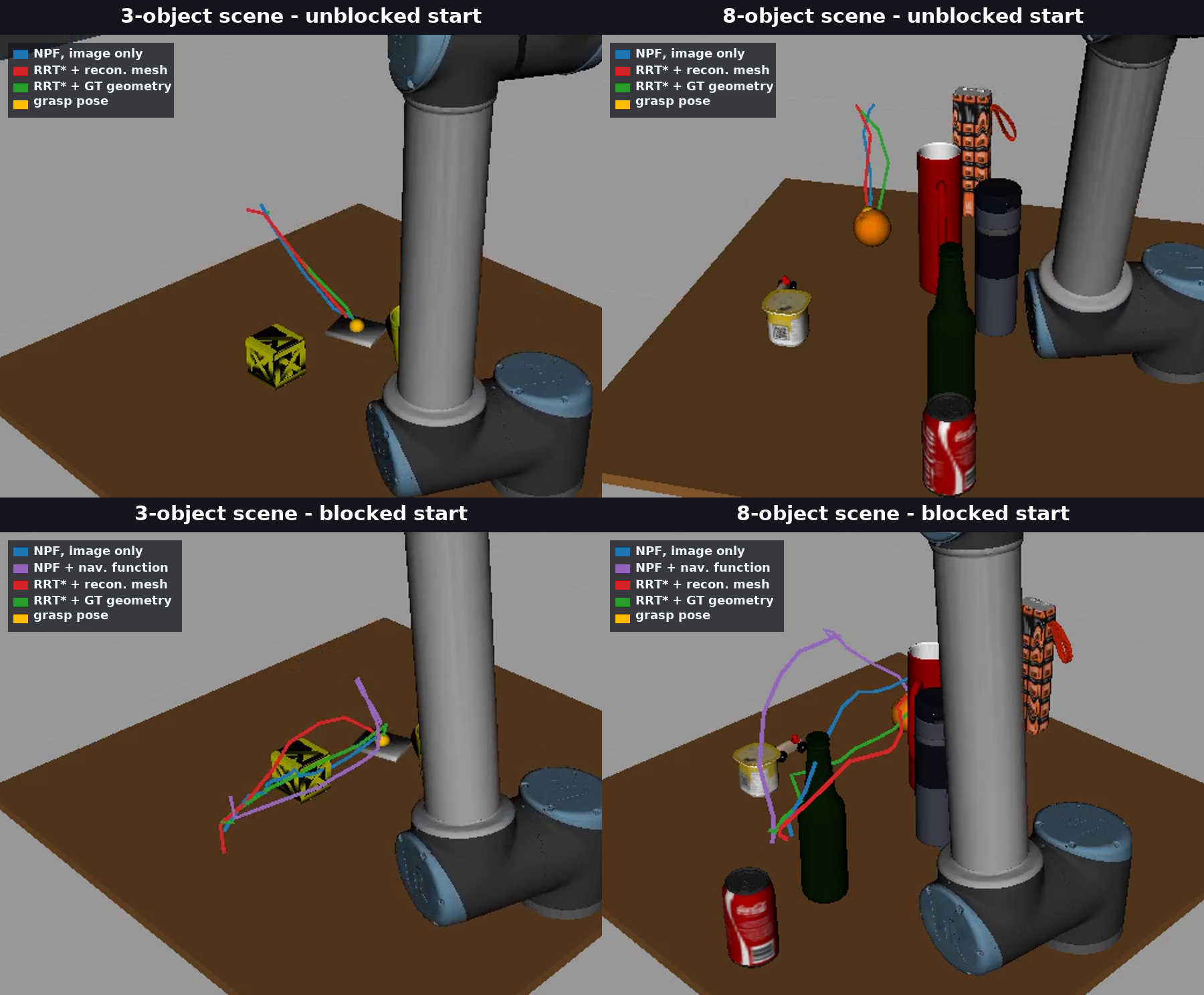}
\caption{Executed trajectories in both scenes, one start per case, rendered in the robot's frame.
Top row: unblocked starts, where a straight line from the start to the grasp is already clear and
every planner reaches it. Bottom row: obstacle-blocked starts, where the image-supervised field
drives at the goal and through the blocking object, while the navigation-function field detours
around it; the two RRT* baselines are shown for reference. Blue: NPF with image supervision only. Purple: NPF with navigation-function
supervision. Red: RRT* on the reconstructed mesh. Green: RRT* on the ground-truth
geometry. Gold: the grasp pose. The four panels share a common field of view.}
% \label{fig:case_tiles}
\label{fig:traj_snapshots}
\end{figure*}
% \begin{figure*}[t]
% \centering
% \begin{subfigure}[b]{0.42\textwidth}
%   \centering\includegraphics[width=\linewidth]{FIGURES/traj_snapshot_book.png}
%   \caption{3 Obj.\ (book) scene, blocked start 7.}
% \end{subfigure}\hfill
% \begin{subfigure}[b]{0.42\textwidth}
%   \centering\includegraphics[width=\linewidth]{FIGURES/traj_snapshot_8obj.png}
%   \caption{8 Obj.\ (orange) scene, blocked start 5.}
% \end{subfigure}
% \caption{Executed TCP trajectories for one obstacle-blocked start per scene (the trials in which all
% four planners succeeded). NPF with navigation-function supervision (blue) detours around or over the
% obstacles with large clearance, whereas the image-supervision-only NPF (light blue) grazes them; RRT*
% on the reconstructed mesh (red) and on the ground-truth geometry (green) are shown for reference. The
% grasp goal is marked in gold.}
% \label{fig:traj_snapshots}
% \end{figure*}

\subsection{Executed Trials in Gazebo}
We execute every planned SE(3) trajectory on a UR10 with a Robotiq three-point gripper in Gazebo
under MoveIt Cartesian control.
Each method is evaluated from the same 10 arbitrary start poses per scene, with varying initial positions and orientations. Fig.~\ref{fig:traj_snapshots} shows representative executed trajectories from one obstacle-blocked start in each scene.
% Fig.~\ref{fig:traj_snapshots} shows one executed trial per scene for all four planners.
% starting from \PoseVFourNpfN{} arbitrary start poses (varied position
% \emph{and} orientation) per scene and reaching the AnyGrasp grasp pose. 
Both methods consume the same
perception: NPF plans directly on the learned field, while RRT* plans on the NeRF-reconstructed mesh
(2\,cm margin, KD-tree over the surface samples) or, as an upper bound, on the ground-truth object
meshes. We record the executed TCP path and the full joint trajectory of every trial so that both
end-effector and full-arm clearance can be audited offline against the ground-truth geometry. In all executed trials the learned repulsive head is left off at inference: with the navigation-function field it adds no clearance and only blocks the grasp approach (Table~\ref{tab:navfn_ablation}, last rows).

\subsection{Training Time and Amortized Cost}
\label{sec:pose_timing}

The one-off cost of each pipeline is NeRF reconstruction of \PoseVFourRecon\,h and \PoseBookRecon\,h against NPF training of \PoseVFourTrain\,h and \PoseBookTrain\,h on one RTX~3090, from the same images (totals \PoseVFourReconTotal\,h vs.\ \PoseVFourNpfTotal\,h and \PoseBookReconTotal\,h vs.\ \PoseBookNpfTotal\,h including pre-processing). Both are paid once per scene; the fair comparison is the amortized time per query, $t_\text{offline}/N+t_\text{plan}$, which NPF wins after a handful of grasps because a plan is a single forward pass (\PoseVFourNpfPlan\,s against \PoseVFourRrtPlan\,s).

\subsection{Failure Modes and Local Minima}
\label{sec:failures}
With image supervision alone the NPF reaches the grasp pose reliably but grazes obstacles: the summed
attractive and repulsive gradients cancel rather than steer around (CF$_\text{GT}$ and Link coll.\ in
Table~\ref{tab:comparison}). The other classical failure of potential fieldsdescent halting at a
spurious minimumdoes not appear in these two scenes (Stall is zero for both variants in
Table~\ref{tab:mesh_comparison_pose}): the wide near-goal sampling radius of the SE(3) losses makes the
attractor effectively global, so the field always reaches the goal, it simply reaches it through the
obstacle rather than around it. The navigation-function supervision of
Sec.~\ref{sec:navfn} addresses exactly this: its target has no minimum other than the goal; the distilled network inherits no such guarantee, so the descent-progress term enforces it empirically by forbidding flat regions in the learned approximation. On these starts it removes the penetrations entirely
(\PoseVFourNpfCFGT\,\% CF$_\text{GT}$, mean clearance 8.8\,cm in the 8-object scene) and drops the
arm-body contacts from \PoseVFourNpfImgLink\,\% to \PoseVFourNpfLink\,\% of executed configurations.
Neither variant stalls on this start setboth fields converge to within 3\,cm of the grasp on every
rollout, and the residual failures are refusals of the Cartesian executor rather than field behaviour.
Scored over all ten starts of each scene rather than only the executed ones, the field reaches the grasp $10/10$ times and penetrates no object: zero collisions, with clearance beyond 7\,cm on $10/10$ paths (8-object).
% and $9/10$ (book), so safety does not depend on which trials the executor ran. 
% Over executed paths it is likewise collision-free on both scenes (\PoseBookNpfExecN/\PoseBookNpfExecN{} book, \PoseVFourNpfExecN/\PoseVFourNpfExecN{} 8-object), while image supervision is collision-free on
% $0/\PoseBookNpfImgExecN$ and $2/\PoseVFourNpfImgExecN$ of its executed paths. 
The book scene, whose starts are laterally blocked by the cup and the crate,
separates the field from the executor: the image-supervised NPF reaches only half of the starts and
every one of its executed paths hits the cup or the crate, whereas the navigation-function NPF
converges within 3\,cm in all ten rollouts and every executed path is collision-free with
13$\times$ the clearance. Its lower executed success rate on this set (40\,\%) is entirely an
execution-layer effect the Cartesian executor aborted four of the converged rollouts before the arm
had moved, and two others executed the full path but finished $2.6$ and $7.0$\,cm from the grasprather than a field failure;
the image-supervised row is bounded the same way: four of its ten rollouts were aborted before motion. The asymmetry with RRT* is in what each planner hands the executor: 7--8 widely spaced waypoints against 26--42 densely specified poses. The refusals persist with collision checking disabled, so they are kinematic, and are not a reorientation effect: the baseline executes even with $129^\circ$ of interpolated rotation.

\subsection{Ablation of the Navigation-Function Supervision}
\label{sec:navfn_ablation}
Table~\ref{tab:navfn_ablation} varies the ingredients of Sec.~\ref{sec:navfn} on the 8-object scene, each variant starting from the same image-trained checkpoint and audited offline from the 10 Gazebo starts (collide: TCP within the 4.5\,cm gripper radius of an obstacle; clear: beyond 7\,cm). The geometry source dominates: targets from the NeRF reconstruction wall off the goal with reconstruction artifacts and no start reaches it, whereas the visual hull recovers every start. The inflation radius trades margin against reach (3.5\,cm never collides but leaves half the paths in the graze band, 10\,cm pinches the approach lanes; 7\,cm is the working point); further distillation rounds convert collisions into clearance, the descent-progress term removes the residual stalls (7/10 to 10/10 reach). 
% and at inference (9/9 reached)

\begin{table}[t]
\caption{Ablation of the navigation-function supervision (8-object scene, offline ground-truth audit,
10 Gazebo start poses). Collide/graze/clear: minimum TCP-obstacle distance $<4.5$\,cm, 4.5-7\,cm,
$\geq 7$\,cm. }
\label{tab:navfn_ablation}
\centering
\footnotesize
\setlength{\tabcolsep}{3.5pt}
\resizebox{\columnwidth}{!}{%
\begin{tabular}{l l r r r r}
\toprule
Factor & Setting & Reach & Collide & Graze & Clear \\
\midrule
\multirow{3}{*}{Inflation radius (3 rounds)} & 3.5\,cm & 10/10 & 0 & 5 & 5 \\
 & 7\,cm & 8/10 & 0 & 4 & 6 \\
 & 10\,cm & 10/10 & 1 & 1 & 8 \\
\midrule
\multirow{3}{*}{Distillation rounds (7\,cm)} & 1 & 10/10 & 2 & 4 & 4 \\
 & 3 & 8/10 & 0 & 4 & 6 \\
 & 8 (model) & 10/10 & 0 & 1 & 9 \\
\midrule
\multirow{2}{*}{Geometry source (7\,cm, 3 rounds)} & visual hull & 8/10 & 0 & 4 & 6 \\
 & NeRF recon.\ mesh & 0/10 & 6 & 2 & 2 \\
\midrule
\multirow{2}{*}{Descent-progress term} & without & 7/10 & 0 & 2 & 8 \\
 & with (model) & 10/10 & 0 & 1 & 9 \\
%  \midrule
% \multirow{1}{*}{Inference barrier (model)$^\dagger$} & - & 9/9 & 0 & 1 & 8 \\
\bottomrule
\end{tabular}}
% \caption{Ablation of the navigation-function supervision (8-object scene, offline ground-truth audit,
% 10 Gazebo start poses). Collide/graze/clear: minimum TCP-obstacle distance $<4.5$\,cm, 4.5-7\,cm,
% $\geq 7$\,cm. }
% \label{tab:navfn_ablation}
\end{table}

\subsection{Loss-Term Ablations}
Table~\ref{tab:loss_ablation} ablates each loss term in the position-only setting (each variant retrained from scratch; 20 fixed and 20 obstacle-blocked starts). The multi-view goal and gradient-alignment terms are structurally necessary without them the learned goal drifts by \AblNoMvGoalErr\,cm or the goal is not a basin while the photometric term is dispensable for planning, supporting the no-reconstruction thesis. The projected-gradient, potential and magnitude terms carry the obstacle information: removing any of them roughly halves the collision-free rate on blocked starts. The anisotropic weight $w_\text{aniso}$ is not justified empirically (removing it \emph{raises} the blocked-start collision-free rate). With matched gradient steps, 8 training views already match 64, and the supervision is robust to segmentation noise: eroding or dilating the masks by 10\,px, or dropping the goal mask in 20\% of views, leaves every metric within noise of the clean model. The remaining variants (view counts, segmentation noise, repulsive decay, near-goal radius, and the 8-object scene) are tabulated in the supplementary material.
\begin{table}[t]
\caption{Loss-term ablation (position-only field, shared terms; each row a single training seed). SR and collision-free rates in \%, on 20 fixed / 20 obstacle-blocked (blk) starts vs.\ ground-truth geometry; Grad.\,q.\ is the validation gradient-alignment score.}
\label{tab:loss_ablation}
\centering
\setlength{\tabcolsep}{4pt}
\resizebox{\columnwidth}{!}{%
\begin{tabular}{l r r r r}
 \toprule
 Variant (3 Obj., position-only) & SR & CF$_\text{GT}$ & CF$_\text{GT}$ blk & Grad.\,q.\\
 \midrule
 Full model & \AblFullSR & \AblFullCFGT & \AblFullBlkCFGT & \AblFullGQ \\
 w/o $\mathcal{L}_c$ (photometric) & \AblNoCSR & \AblNoCCFGT & \AblNoCBlkCFGT & \AblNoCGQ \\
 w/o $\mathcal{L}_g$ (gradient field) & \AblNoGSR & \AblNoGCFGT & \AblNoGBlkCFGT & \AblNoGGQ \\
 w/o $\mathcal{L}_\text{pot}$ & \AblNoPotSR & \AblNoPotCFGT & \AblNoPotBlkCFGT & \AblNoPotGQ \\
 w/o $\mathcal{L}_\text{mag}$ & \AblNoMagSR & \AblNoMagCFGT & \AblNoMagBlkCFGT & \AblNoMagGQ \\
 w/o $\mathcal{L}_\text{align}$ & \AblNoAlignSR & \AblNoAlignCFGT & \AblNoAlignBlkCFGT & \AblNoAlignGQ \\
 w/o $\mathcal{L}_\text{mv}$ & \AblNoMvSR & \AblNoMvCFGT & \AblNoMvBlkCFGT & \AblNoMvGQ \\
 w/o $w_\text{aniso}$ & \AblNoAnisoSR & \AblNoAnisoCFGT & \AblNoAnisoBlkCFGT & \AblNoAnisoGQ \\
 \bottomrule
\end{tabular}}
% \caption{Loss-term ablation (position-only field, shared terms; each row a single training seed). SR and collision-free rates in \%, on 20 fixed / 20 obstacle-blocked (blk) starts vs.\ ground-truth geometry; Grad.\,q.\ is the validation gradient-alignment score.}
% \label{tab:loss_ablation}
\vspace{-3mm}
\end{table}

\subsection{Qualitative Analysis}

% \begin{figure}[t]
%     \centering
%     \includegraphics[width=\columnwidth]{FIGURES/Demo3_real_robot_only.png}
%     \caption{Rollouts in simulation (top) and on the real robot (middle, bottom): the arm reaches the target while avoiding the other objects.}
%     \label{fig:robot_exp}
%     \vspace{-3mm}
% \end{figure}

\begin{figure}[t]
    \centering
    \includegraphics[width=\columnwidth]{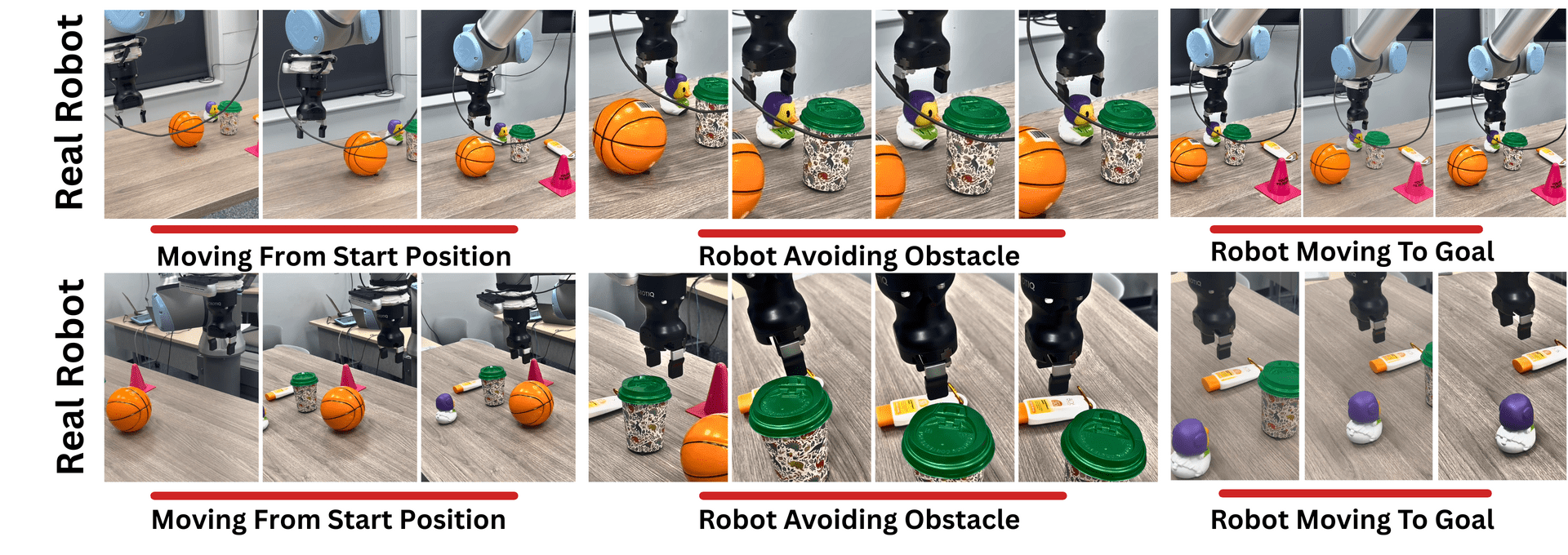}
    \caption{Real-robot rollouts of the \emph{position-only} field, to two different targets: the arm leaves the start pose, avoids the intervening objects, and reaches the goal. Hardware runs use the position-only formulation; the SE(3) formulation of this paper is evaluated in executed simulation trials (Tables~\ref{tab:comparison}--\ref{tab:mesh_comparison_pose}). These rollouts are demonstrations and are reported as such (Sec.~\ref{sec:failures}).}
    \label{fig:robot_exp}
    \vspace{-3mm}
\end{figure}

% Fig.~\ref{fig:robot_exp} shows rollouts from start to grasp in simulation and on the real robot; because the field is followed continuously rather than sampled, the executed paths are spatially smoother than the RRT* paths of Fig.~\ref{fig:traj_snapshots}.

Fig.~\ref{fig:robot_exp} shows rollouts from start to grasp on the real robot, using the position-only field; because the field is followed continuously rather than sampled, the executed paths are spatially smoother than the RRT* paths of Fig.~\ref{fig:traj_snapshots}. These hardware runs are demonstrations of the earlier position-only formulation: the SE(3) formulation of this paper is evaluated in the executed simulation trials reported above, and quantitative hardware trials remain future work.

\section{Discussion and Limitations}
Reconstruct-then-plan pipelines commit early to a hard geometric representation and pay for it in reconstruction time (\PoseVFourRecon\,h and \PoseBookRecon\,h of NeRF training here) and in planner queries that must check a dense model. Our field is trained once from the same images, poorly observed regions contribute weak, smooth gradients rather than a possibly wrong surface, and a plan is a sequence of network evaluations; the navigation-function supervision keeps this architecture, running the planner once offline on image-derived geometry and evaluating only the distilled field at run time. Limitations: executed success is bounded by the Cartesian executor rather than by the field (Sec.~\ref{sec:failures}); execution is slower than RRT* because of conservative gradient steps; the target pose and the visual hull inherit the errors of AnyGrasp and of the instance detector; and performance depends on a scene's training distribution, as for reconstruction systems. Adaptive step sizing, learned velocity control and dynamic scenes are natural next steps.

\section{Conclusion}
We presented an SE(3) neural potential field learned directly from images, whose gradient plans 6-DoF grasp reaches without explicit 3D reconstruction, and showed how supervising it with a navigation function computed on image-derived geometry removes the spurious minima and grazing of classical potential fields. On executed Gazebo trials the navigation-function NPF is collision-free against the ground-truth geometry in every executed path of both scenes (from 25\,\% and 0\,\% with image supervision alone), keeps 8.6-8.8\,cm of clearance, and plans in about 2\,s against 67-133\,s for RRT* on the reconstruction, a margin set by dense collision checking rather than planner complexity (Table~\ref{tab:mesh_comparison_pose}).

\bibliographystyle{IEEEtran}
\bibliography{references}

\end{document}